\pdfoutput=1

\documentclass[11pt]{article}

\usepackage[]{acl}

\usepackage{times}
\usepackage{latexsym}
\usepackage{amsmath}
\usepackage{amssymb}
\usepackage{booktabs}
\usepackage{xcolor,colortbl}
\usepackage{listings}
\usepackage[most]{tcolorbox}
\usepackage{caption}
\usepackage{longtable}
\usepackage{pifont}
\usepackage{enumitem}

\newtcolorbox{promptbox}{
  colback=gray!10,
  colframe=gray!60,
  arc=4mm,
  boxrule=0.5pt,
  width=\linewidth,
  boxsep=8pt
}

\usepackage[T1]{fontenc}

\usepackage[utf8]{inputenc}

\usepackage{microtype}

\usepackage{inconsolata}

\usepackage{graphicx}

\title{Simulated Students in Tutoring Dialogues: Substance or Illusion?}

\author{
 \textbf{Alexander Scarlatos\textsuperscript{1}},
 \textbf{Jaewook Lee\textsuperscript{1}},
 \textbf{Simon Woodhead\textsuperscript{2}},
 \textbf{Andrew Lan\textsuperscript{1}}
\\
 \textsuperscript{1}University of Massachusetts Amherst,
 \textsuperscript{2}Eedi
\\
 \small{
   \texttt{\{ajscarlatos,jaewooklee,andrewlan\}@cs.umass.edu}
   }, \small{
   \texttt{simon.woodhead@eedi.co.uk}
 }
}

\begin{document}
\maketitle
\begin{abstract}
Advances in large language models (LLMs) enable many new innovations in education. However, evaluating the effectiveness of new technology requires real students, which is time-consuming and hard to scale up. Therefore, many recent works on LLM-powered \emph{tutoring} solutions have used \emph{simulated students} for both training and evaluation, often via simple prompting. Surprisingly, little work has been done to ensure or even measure the quality of simulated students. In this work, we formally define the student simulation task, propose a set of evaluation metrics that span linguistic, behavioral, and cognitive aspects, and benchmark a wide range of student simulation methods on these metrics. We experiment on a real-world math tutoring dialogue dataset, where both automated and human evaluation results show that prompting strategies for student simulation perform poorly; supervised fine-tuning and preference optimization yield much better but still limited performance, motivating future work on this challenging task.\footnote{Our code and data annotations are available at \url{https://github.com/umass-ml4ed/sim-student-eval}}

\end{abstract}

\section{Introduction}

Private tutoring has been shown to be highly effective for helping students learn. Naturally, with the emergence of powerful large language models (LLMs), significant attention has been put on how to effectively use LLMs as tutors. Many LLM-based tutors have already been deployed by learning platforms and companies \cite{khanmigo,livehint,studymode,learnlm,claudeedu}, aiming at scaling up the impact of on-demand, real-time tutoring. 

Efforts in developing LLM-based tutors can be broadly categorized into two types: alignment-based and student-based. The first type relies on aligning LLMs with well-established pedagogical principles, especially on how to respond to student utterances during tutoring dialogues to maximize learning and engagement~\cite{khanmigo,scarlatos2025training,sonkar2024pedagogical}.
The second type resorts to training, often via reinforcement learning (RL), with students in-the-loop; since doing RL training with real human students at large scale is difficult, existing works either ask experts to role-play students~\cite{learnlm} or prompt LLMs to simulate students~\cite{eth-tutor-rl,2024.EDM-long-papers.6,litype,mathdial,zhang2024simulating}. 

However, both of these student simulation approaches have obvious limitations: the former is still not scalable, while the latter requires LLM-based students to behave realistically. Many recent works have shown that LLMs cannot reliably simulate student behavior simply through prompting \cite{eth-sim-student,sonkar2023class}.
Moreover, it is challenging to make LLM-based simulated students follow cognitive characteristics of real human students, like the power law of practice \cite{dialogue-kt,tutorgym}, when learning.
In addition to these well-known shortcomings, there do not exist many evaluation metrics for realistically simulating students in dialogues, particularly at the turn-level; recent works focus on linguistic features at the population-level \cite{perczel2025teachlm}, or consistency with synthetic profiles at the dialogue-level \cite{liu-etal-2024-personality}. However, it is difficult to determine if systems developed using simulated students are reliable without thoroughly verifying the simulated students themselves.
We discuss prior work in more detail in Appendix~\ref{sec:rw}.

\paragraph{Contributions}
In this paper, we explore the task of creating more realistic LLM-based simulated students using a two-stage approach. First, we identify six high-level dimensions, identified by the learning sciences research community, that define student behavior in dialogues: 1) dialogue acts, 2) correctness, 3) error-making, 4) knowledge acquisition, 5) language use, and 6) tutors' responses. For each dimension, we develop reference-based automated evaluations to measure the realism and faithfulness of a simulated student response with respect to a ground-truth response. Second, we benchmark performance on a set of strong simulated student methods, 
including several prompting-based methods, supervised fine-tuning (SFT), and one using multi-objective RL. To the best of our knowledge, our work is the first to 1) develop reference-based metrics for realistic student simulation in dialogues, and 2) examine the ability of LLMs to replicate real student turns in dialogues. 
We conduct extensive experiments on a real-world dataset with 2,000 dialogues between human tutors and students.
Using both automated metrics and human evaluations, we find that: (various styles of) prompting often fail to capture most dimensions of student behavior, SFT significantly improves the alignment of generated student utterances with these dimensions, and RL leads to further improvements. We further show that our automated metrics show strong agreement with human experts.

\section{Methodology}

We now detail our notations, methodology for simulating students in dialogues, and evaluating the faithfulness of these simulations. A dialogue $d=(s_0,t_1,s_1,\ldots,t_M,s_M)$ is defined as an alternating sequence of student turns, $s_i$, and tutor turns, $t_i$, where $M$ is the number of \textit{turn pairs} in the dialogue; $s_0$ is present if the student initiates the dialogue, and $s_M$ is present if the student ends the dialogue. For a dataset of $N$ real tutor-student dialogues, $\mathcal{D}$, $d^n$ denotes the $n$-th dialogue. We refer to the textual content of a turn as an ``utterance''. Dialogues are often grounded in a question, $q$, that the student is attempting and the tutor is guiding them through. In these cases, the question is associated with a set of \textit{knowledge components (KCs)}, $C$, i.e., skills that the student must possess to answer the question correctly.

\subsection{LLM-based Student Simulation}
\label{sec:student-modeling}

We simulate realistic student behavior by using an LLM, $\mathcal{M}_S$, to generate a predicted student utterance, $\hat{s}_i$, at each turn $i$ in the dialogue, conditioned on the current dialogue history and any other relevant context. Formally, this process is summarized as $\hat{s}_i\sim\mathcal{M}_S(s|s_{<i},t_{\le i},q,P)$. $P$ represents a prompt, which includes instructions on simulating student behavior, but can also include additional context on the student, such as a \textit{persona}, if available. $\mathcal{M}_S$ can be a pre-trained LLM, but can also be fine-tuned using SFT on a dataset of existing student utterances, minimizing the negative log likelihood of student utterances, i.e., $\mathcal{L} = -\sum_{n=0}^N \sum_{i=0}^M \log \mathcal{M}_S(s^n_i|s^n_{<i},t^n_{\le i},q^n,P^n)$. We can further refine $\mathcal{M}_S$ using RL in order to optimize for specific properties of student behavior, which we detail later.

\subsection{Evaluating Simulated Student Turns}
\label{sec:eval_sim_student}

\begin{figure*}
    \centering
    \includegraphics[width=.95\linewidth, trim=1 1 1 1, clip]{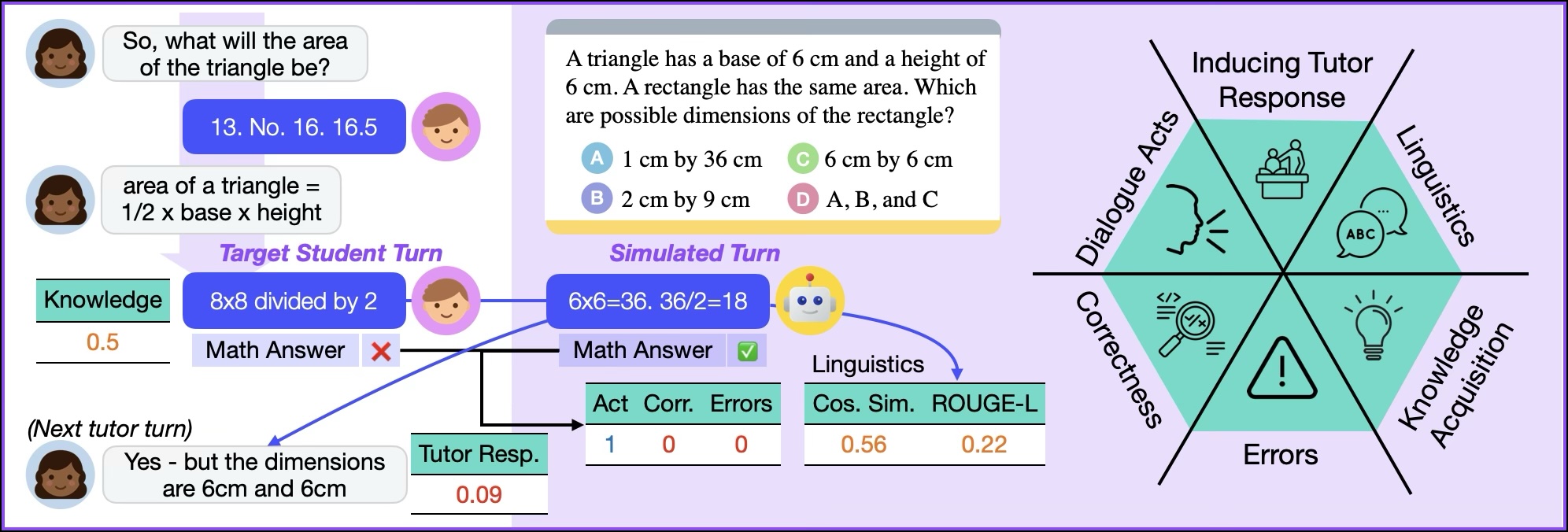}
    \caption{Overview of the seven evaluation metrics for simulated student turn evaluation, with a real paraphrased tutor–student dialogue serving as the reference. In this example, the ground-truth student turn and the simulated turn have the same dialogue act \textit{Math Answer}; however, the target turn is incorrect while the simulated turn is correct.}
    \label{fig:eedi_ex}
\end{figure*}

We ground our evaluation metrics in several high-level measurable properties of student utterances across six dimensions. These metrics cover important aspects of students, such as behavior, knowledge, and language use, that enable us to make meaningful comparisons between real and simulated students. 
Several of the metrics rely on ground-truth labels for real student turns; we use LLM annotation to provide these labels, via GPT-4.1~\cite{gpt-4.1}. We provide these prompts in Appendix~\ref{sec:prompts} and also show the agreement between the labels and human experts in Table~\ref{tab:turn-agreement}. Finally, several metrics require fine-tuned LMs, as detailed below. We provide the prompts for these models in Appendix~\ref{sec:prompts} and additional implementation details in Appendix~\ref{sec:implementation-details}.

\paragraph{Dialogue Acts}

Our first student behavior measure is dialogue acts, a widely studied approach for classifying actions taken in dialogues. We develop a list of five student dialogue acts in Table~\ref{tab:acts}, rooted in prior work \cite{dialogue-acts, 10.1007/978-3-031-98417-4_9, studychat}. We define $a_i$ as the act that the student takes in turn $s_i$ and $\hat{a}_i$ as the act in the simulated student turn $\hat{s}_i$. We define the similarity to be binary-valued, i.e., $1$ if $a_i=\hat{a}_i$ and $0$ otherwise. We prompt an LLM to provide the ground-truth act labels. To reduce the costs of our evaluation metric, we then fine-tune a local LLM, $\mathcal{M}_\text{act}$, to classify dialogue acts, trained on the ground-truth act labels. At evaluation time, we use this model to classify the acts of simulated student turns, i.e., $\hat{a}_i \sim \mathcal{M}_\text{act}(a|\hat{s}_i,s_{<i}, t_{\le i})$.

\paragraph{Correctness}

Another important aspect of student behavior is whether they are responding correctly to questions (or tasks) raised by the tutor during the dialogue, who often ``scaffold'' problem solving by asking small sub-questions \cite{azevedo2005scaffolding}. 
Concretely, we define $y_i$ as the correctness of $s_i$; $y_i=1$ if $s_i$ is a correct response to the question in the tutor turn $t_i$, and $y_i=0$ if it is an incorrect response. Additionally, $y_i=na$ if the tutor does not ask a question in $t_i$; these turns include off-topic ones or simple utterances that keep the dialogue flowing. Similarly, $\hat{y}_i$ indicates the correctness of the simulated student turn, $\hat{s}_i$. When $y_i\ne na$, we consider the correctness similarity to be 1 if $y_i=\hat{y}_i$ and 0 otherwise. When $y_i=na$, we do not measure correctness similarity. 
We use an LLM to provide ground-truth correctness labels.
We find that classifying correctness by fine-tuning an LLM is much more challenging than doing so for dialogue acts. 
Instead, we prompt a reasoning-based LLM, GPT-5 mini~\cite{gpt-5}, with low reasoning effort, to evaluate the correctness of simulated student turns, including the ground-truth turn as reference. 

\paragraph{Errors}

In addition to correctness, it is also important to understand the \textit{errors} that students make when they give incorrect answers \cite{eedi-mining-misconceptions-in-mathematics}.
We define $e_i$ as the error in the ground-truth student turn and $\hat{e}_i$ as the error in the simulated student turn. If the ground-truth turn has an error, we set the error similarity to 1 if $e_i=\hat{e}_i$ and 0 otherwise. We do not evaluate similarity if the ground-truth turn is not incorrect. In practice, we only need to test error equivalence, avoiding the challenge of classifying errors.
Therefore, we prompt GPT-5 mini to determine if two turns have the same error. 
We perform correctness and error evaluation in a single prompt to reduce costs.

\paragraph{Knowledge Acquisition}
In order for simulated students to be useful for modeling learning outcomes, they need to follow \textit{knowledge acquisition} patterns that are similar to real students \cite{tutorgym}. 
We formalize knowledge acquisition using knowledge tracing (KT) \cite{kt}, where knowledge is represented as mastery of KCs. 
Following \citet{dialogue-kt}, a framework for KT in tutoring dialogues, we use an LLM-based model to predict student correctness in subsequent turns. Formally, we define $C_i \subseteq C$ as the set of KCs relevant to the tutor turn $t_i$, which we identify using LLM prompting.

Using these KC labels and the correctness labels, as detailed above, we train a model, $\mathcal{M}_\text{KT}$, to estimate a student's current \textit{knowledge state}, which represents the student's current mastery over all KCs in $C$. We achieve this by training the model to predict if the student will correctly answer the \textit{next} tutor-posed task correctly, conditioned on the dialogue so far. 
Formally, the estimated probability of a correct answer is $\frac{1}{|C_{i+1}|} \sum_{k=1}^{|C_{i+1}|} z_{i+1,k}$, where $z_{i+1,k} = \mathcal{M}_\text{KT}(c_{i+1,k}|s_{\le i}, t_{\le i})$ represents the student's estimated mastery of the KC $c_{i+1,k}$.

To measure the \textit{acquisition} of knowledge, we define $\nabla Z_i = \{z_{i,1} - z_{i-1,1},\ldots,z_{i,|C|} - z_{i-1,|C|}\}$, i.e., how much the student's mastery changes for each KC since the last turn. We similarly compute the change in knowledge for the simulated student turn, $\nabla \hat{Z}_i = \{\hat{z}_{i,1} - z_{i-1,1},\ldots,\hat{z}_{i,|C|} - z_{i-1,|C|}\}$, where $\hat{z}_{i+1,k} = \mathcal{M}_\text{KT}(c_{i+1,k}|\hat{s}_i, s_{< i}, t_{\le i})$. 
Since the range of these values can be highly dependent on $\mathcal{M}_\text{KT}$ and the data it is trained on, we compare the quantile buckets of mastery deltas rather than the raw values themselves. Specifically, we compute 5 equal sized mass-based quantile bins for all $\nabla Z_i$, where $\operatorname{quant}(z) \in \{0,\ldots,4\}$ returns the quantile index. We then define the similarity between $\nabla Z_i$ and $\nabla \hat{Z}_i$ as $1 - \sum_{k=1}^{|C|}\frac{|\operatorname{quant}(\nabla Z_{i,k})-\operatorname{quant}(\nabla \hat{Z}_{i,k})|}{4|C|} \in [0,1]$, i.e., the inverted average distance between the delta quantiles. We provide a step-by-step computation of knowledge acquisition similarity in Table~\ref{tab:knowledge-breakdown} to illustrate the intuition for this metric.

\paragraph{Linguistics}
Prior works have found that LLM-based simulated students often fail to reproduce basic linguistic patterns of real students \cite{eth-sim-student}. 
Therefore, we employ existing text-based similarity metrics, specifically the cosine similarity between the text embeddings of the simulated and ground-truth utterances, which measures their semantic similarity, and ROUGE-L \cite{lin-2004-rouge}, which measures word-level recall. Both of these measures capture many aspects of linguistic patterns in student utterances, such as specific word use, utterance length, and sentiment. We use the \texttt{Qwen3-Embedding-8B} model \cite{qwen3embedding} in the cosine similarity calculation.

\paragraph{Inducing Tutor Responses} Finally, we measure how well student turns fit into the natural progression of a dialogue. We can examine if the simulated student response is likely to induce the actual response from the tutor, assuming that the tutor would respond similarly if the predicted student utterance is similar to the actual one. 
Therefore, we leverage the likelihood of the \emph{ground-truth, next tutor turn} conditioned on the simulated student utterance in this turn. We fine-tune an LLM tutor model, $\mathcal{M}_T$, with SFT on the ground-truth tutor utterances and calculate the inverse perplexity of the tutor turn. Formally, this metric is defined as $\exp\big(-\frac{1}{|t_{i+1}|}\log P_{\mathcal{M}_{T}}(t_{i+1}|\hat{s}_i,s_{<i},t_{\le i}) \big)$.

\subsection{Tuning with RL}
\label{sec:rl_tuning}

Given these automated evaluation metrics, we can create a \textit{reward function} to provide real-time feedback on simulated student responses to improve them via RL. 
Specifically, we take the average of all metrics to form our final reward, which is possible, since all metrics are in the $[0,1]$ range; this simple aggregation can form a Pareto-optimal policy on convex Pareto fronts \cite{lin2019pareto}, and we leave more complex aggregations for future work. In this work, we use offline RL rather than online algorithms to reduce computational burden.

We use a simple four-stage pipeline for offline RL training: we 1) generate $n$ candidate student responses for each dialogue turn in the train set using an SFT-ed student model, 2) evaluate each candidate student response on each of our evaluation metrics, 3) form preference pairs between all possible pairs of the $n$ candidate responses for each turn, where a response is preferred if it has a higher average score across metrics, and 4) train the student model on the resulting preference pairs using direct preference optimization (DPO) \cite{rafailov2023direct}. In practice, we employ two techniques for eliminating noisy preference pairs. First, following \cite{smart}, we only form a pair if the score difference is greater than a threshold, $\epsilon$. Second, we do not train on any of the first 5 turns in a dialogue, since we find that student behavior is highly uncertain early in dialogues, resulting in noisy reward signals.

We note that the purpose of our RL pipeline is to explore the potential of leveraging our metrics to improve simulated student realism. However, we caution that training on LLM-evaluated metrics can lead to reward hacking, where the trained model can learn to overfit to errors in the evaluations, potentially causing them to become unreliable. For this reason, we encourage future works to carry out expert human evaluation in situations when evaluation metrics are leveraged during training.

\section{Experiments}
We now detail our experimental setup to 1) benchmark a variety of student simulation methods on a real-world dataset and 2) validate our automated evaluation metrics using human evaluation.

\subsection{Dataset}

We conduct our experiments on Question-Anchored Tutoring Dialogues 2k \cite{eedi-dialogues}, the largest publicly-available dataset of real student-tutor dialogues from the \citet{eedi_labs} learning platform. Each dialogue centers around a middle school student solving a math multiple-choice problem, where a trained tutor guides the student through the process via online chat. Problems span a wide range of topics, including Algebra, Geometry, and Number Sense. After processing, the dataset contains 1,529/382 dialogues in the pre-defined train/test split, respectively. We further split the train set into a train/validation set with 1,147/382 dialogues, respectively. We train all models on the train set, use the validation set for hyperparameter tuning, and show results on the test set. Additional details are available in Appendix~\ref{sec:dataset-details}.

\subsection{Simulated Student Benchmarks}
\label{sec:benchmarks}

We now detail the methods that we benchmark for the student simulation task, including fine-tuning and prompting approaches.

\subsubsection{Fine-Tuning}

We use both \texttt{Llama-3.1-8B-Instruct} and \texttt{Llama-3.2-3B-Instruct} \cite{llama} to examine the impact of model size, and train all models using LoRA~\cite{hu2022lora}. We discuss training and inference details in Appendix~\ref{sec:model_details} and show prompts in Appendix~\ref{sec:prompts}.
We perform standard \textbf{SFT} on all student turns in the train set, as detailed above in Section~\ref{sec:student-modeling}.
We then further perform \textbf{DPO} training as detailed above in Section~\ref{sec:rl_tuning}. 

\subsubsection{Prompting}

One of the most common approaches for simulating students is to prompt pre-trained LLMs to behave like real students. Therefore, we test a variety of prompting approaches to see how they compare to fine-tuning. Unless otherwise specified, we prompt GPT-4.1 for each of the following methods. We provide prompts, including persona annotation prompts, in Appendix~\ref{sec:prompts}, and model decoding details in Appendix~\ref{sec:model_details}. We also provide an example persona, summary, and dialogue retrieval in Table~\ref{tab:persona-examples}.

\paragraph{Zero-Shot}
We simply instruct the LLM to behave like a real student with a few simple guidelines for student behavior, adapting the prompt from \citet{eth-tutor-rl} with minimal changes.

\paragraph{OCEAN Persona}
In prompting, we can provide a \textit{persona} to the LLM, instructing it to behave like a student while following the traits described in the persona \cite{2024.EDM-long-papers.6}. We use OCEAN (i.e., ``Big Five'') personas \cite{mccrae1992introduction}, which are commonly used in human user simulation \cite{liu-etal-2024-personality, kim2025propersimdevelopingproactivepersonalized}. Specifically, we first prompt an LLM to estimate each of the five OCEAN traits as ``low'', ``neutral'', or ``high'' given the full, ground-truth dialogue. We then provide this estimated persona in the simulated student prompt. We note that this approach does leak information from the ground-truth. If prior student dialogues are available, we can use them to estimate the persona; however, they are not available in the dataset we are using.

\paragraph{Oracle}
As an upper bound for prompting-based methods, we investigate how well prompted models can simulate student behavior given a summary of the current dialogue.
Specifically, we first have an LLM summarize the student's behavior in the dialogue, including OCEAN traits and learning patterns. We then provide this summary in the simulated student prompt, revealing the student's behavior ahead of time.

\paragraph{In-Context Learning (ICL)}
We also test ICL where an example dialogue is given in the prompt. Adopting the approach in \cite{lee-etal-2024-effective}, we use dialogue summaries to retrieve examples. We first encode the Oracle summaries using  \texttt{Qwen3-Embedding-8B} and then select the dialogue in the training set whose summary is closest to the summary of the current dialogue to be evaluated. Similar to OCEAN, this approach technically leaks information from the ground-truth and prior dialogues from the same student should be used, if available.

\paragraph{Reasoning}
Finally, we experiment with a reasoning LLM, GPT-5 mini, with medium reasoning effort. In addition to instructions from the Zero-Shot prompt, we include descriptions of the evaluation criteria that the response will be judged on, letting the model reason about how to generate student-like utterances.

\subsection{Evaluation Metrics}
\label{sec:metrics}

We use seven total metrics to evaluate simulated student dialogue turns, as discussed in Section~\ref{sec:eval_sim_student}: \textbf{Acts} for dialogue acts, \textbf{Corr.} and \textbf{Errors} for the correctness and errors in student utterances, \textbf{Knowledge} for knowledge acquisition patterns, \textbf{Cos. Sim.} (cosine similarity) and \textbf{ROUGE-L} for linguistic patterns, and \textbf{Tutor Resp.} for the likelihood of inducing tutor responses.

\begin{table*}[ht]
    \centering
    \small
    \begin{tabular}{lccccccc}
        \toprule
        Method & Acts & Corr. & Errors & Knowledge & Cos. Sim. & ROUGE-L & Tutor Resp. \\
        \midrule
        \rowcolor{gray!21} \multicolumn{8}{c}{Fine-Tuning Methods}\\
        \midrule
        SFT (Llama 3.2 3B) & 0.6645 & 0.5557 & \underline{0.0815} & 0.8726 & 0.7331 & 0.3058 & 0.2033 \\
        DPO (Llama 3.2 3B) & \underline{0.6762} & 0.5748 & 0.0584 & 0.8745 & 0.7345 & \underline{0.3109} & 0.2037 \\
        \midrule
        SFT (Llama 3.1 8B) & 0.6671 & 0.5670 & 0.0661 & \underline{0.8766} & \underline{0.7383} & \textbf{0.3212} & \underline{0.2038} \\
        DPO (Llama 3.1 8B) & \textbf{0.6840} & 0.5761 & 0.0529 & \textbf{0.8787} & \textbf{0.7390} & \textbf{0.3212} & \textbf{0.2039} \\
        \midrule
        \rowcolor{gray!21} \multicolumn{8}{c}{Prompting Methods}\\
        \midrule
        Zero-Shot (GPT 4.1) & 0.4998 & 0.5926 & 0.0220 & 0.8078 & 0.5460 & 0.1648 & 0.1911 \\
        OCEAN  (GPT 4.1) & 0.5268 & \underline{0.6039} & 0.0308 & 0.8135 & 0.5739 & 0.1772 & 0.1942 \\
        ICL (GPT 4.1) & 0.5085 & 0.5991 & 0.0319 & 0.8138 & 0.5919 & 0.1939 & 0.1914 \\
        Reasoning (GPT 5 Mini) & 0.5755 & 0.5870 & 0.0088 & 0.8395 & 0.5992 & 0.2170 & 0.1909 \\
        \midrule
        Oracle (GPT 4.1) & 0.5097 & \textbf{0.6755} & \textbf{0.1872} & 0.8063 & 0.6032 & 0.2109 & 0.1942 \\
        \bottomrule
    \end{tabular}
    \caption{Results on turn-level student utterance prediction. Prompting-based methods perform well on correctness, while fine-tuned methods perform well on acts, knowledge, linguistic similarity, and tutor responses. All methods, other than Oracle, perform poorly on error prediction. Best method is \textbf{bolded} and second best is \underline{underlined}.}
    \label{tab:turn-results-main}
\end{table*}

\subsection{Human Evaluation}

To gauge the validity of our dialogue turn-level student simulation performance metrics, we conduct an IRB-approved human evaluation with three evaluators experienced in math teaching or tutoring, recruited from Upwork\footnote{\url{https://www.upwork.com/}}.
In total, we collect annotations on 190 turns across 38 dialogues, with 20 turns shared across two evaluators for inter-rater agreement. See Appendix~\ref{sec:human_eval_details} for additional details.

For each dialogue, an evaluator examines five consecutive student turns in a dialogue, first for ground-truth responses and then for simulated ones. We include simulated responses from DPO trained on the 8B model, Zero-Shot, and Oracle, which cover a wide range of performance levels across the metrics. We inform the evaluator whether a turn is ground-truth, and randomly shuffle the order of simulated turns to not reveal which method produced which response. We collect annotations after the fifth turn in each dialogue to provide sufficient context.
For both ground-truth and simulated turns, we ask evaluators to label the dialogue act and correctness. For simulated turns that they label as incorrect, we also ask them to evaluate if they have the same error as the corresponding ground-truth turn. Finally, for simulated turns, we ask evaluators to rate linguistic similarity to the ground-truth on a 5-point Likert scale. We compute 1) the human-evaluated scores for acts, correctness, errors and linguistic similarity, 2) agreement between these four categories and the corresponding automated metrics, 3) agreement between human labels and LLM-assigned labels for acts and correctness on ground-truth turns, and 4) agreement between evaluators on the shared set of turns.

\section{Results}

We now detail our experimental results, including quantitative results through automated and human evaluation, a qualitative analysis of simulated student utterances, and finally an ablation study.

\subsection{Quantitative Results}

Table~\ref{tab:turn-results-main} shows quantitative results for turn-level student simulation. We see that fine-tuning methods generally outperform prompting methods, with significant improvements on the Acts, Knowledge, Cos.\ Sim., ROUGE-L, and Tutor Resp. metrics. They also perform better on Errors, with the exception of Oracle, which contains leaked information on exact errors made by students in its prompt. While prompting methods perform better on the correctness metric, this can be attributed to mostly generating correct responses, the majority class, as seen in Figure~\ref{fig:correctness-distribution}. These trends indicate that while LLM prompting can anticipate some high-level behavior of the student, fine-tuning is required to capture more nuanced details.

Different prompting-based methods have clearly different strengths and weaknesses. As expected, Zero-Shot performs the worst, showing the importance of context in the prompt. Between OCEAN and ICL, OCEAN performs better on Acts, which can be explained by the OCEAN persona containing high-level behavioral traits for the student. ICL performs better on linguistic metrics, which can be explained by the model reflecting student language patterns in the example dialogues. 
The Reasoning method performs significantly better than other prompting approaches on several metrics, including Acts, Knowledge, and ROUGE-L. This result shows that reasoning clearly helps, although less so than fine-tuning. Finally, while Oracle outperforms all other methods on Correctness and Errors due to information leakage in the summaries, it does not perform very well on other metrics. This result shows that LLM prompting is highly limited for student simulation; even with a ``cheat sheet'' in the prompt, the model cannot outperform much smaller, fine-tuned models on most metrics.

Perhaps surprisingly, we find that DPO only slightly outperforms SFT on all metrics, with even slightly worse performance on Errors, which is possibly due to a sparse reward signal: since candidate turns are sampled from the SFT model, which does not perform well on errors either, there are not enough positive samples in the data for RL training.
Similarly, we find that the larger 8B model outperforms the smaller 3B model by small but consistent margins on all metrics except errors. We postulate that this result is due to the inherent difficulty of predicting what students will do next, especially in open-ended settings like dialogues. More advanced techniques, such as reasoning, online RL, or data augmentation, may be needed to further improve the performance of RL, which are possible directions for future work.

\subsection{Human Evaluation}

Table~\ref{tab:turn-results-human-eval} shows the performance of different simulated student methods using human-evaluated proxies of our evaluation metrics. Mostly, results match the patterns on automated metrics: DPO performs best on Acts and Linguistic, while Oracle performs best on Correctness and Errors. Across Acts, Correctness, and Errors, the values tend to be higher than on automated metrics, though this result is likely due to human evaluation being performed on later dialogue turns; earlier turns are more challenging, as seen in Figure~\ref{fig:results-turn-level}. DPO also performs slightly better than Zero-Shot on Correctness, whereas on the automated metrics it is slightly worse. The likely cause is that annotators are more likely to label Zero-Shot responses as $na$, possibly due to their verbosity. Overall, these results confirm that fine-tuning leads to more realistic student simulations than prompting, although having an Oracle in the prompt helps. However, we note that even the best-performing methods are poor, although it is currently unclear what the theoretical upper bound for performance on our metrics is (the upper bound should be less than 1 on each metric due to inherent randomness in student behavior). Therefore, it is important to study more advanced techniques for realistically simulating students and predicting student behavior in future work.

\paragraph{Metric and Label Agreement}

Table~\ref{tab:turn-agreement} shows agreement-based reliability results for automated evaluation, LLM annotation, and human evaluation. For Acts, Correctness, and Errors, we report Cohen's Kappa, and for Linguistic, we report Pearson's correlation coefficient. The agreement between human-assigned scores and our automated metrics is very high across labels, indicating that our automated metrics give a reliable measure of student simulation quality. The Errors agreement is slightly lower because 1) it is only computed on incorrect turns, resulting in a smaller sample size, and 2) labels are highly imbalanced, with most simulated turns not making the same error as the ground-truth turn. There is also substantial agreement on the ground-truth turns between human-assigned labels and LLM-assigned labels for Acts and Correctness, showing that our prompting-based method for this annotation is reliable. Finally, we find that inter-rater agreement is very high for Correctness, Errors, and Linguistic. The moderate agreement for acts is primarily due to a single label, Math Answer, being selected much more frequently than the other labels (71- 83\% across annotators).

\begin{table}[h]
    \centering
    \small
    \begin{tabular}{lcccc}
        \toprule
        Method & Acts & Corr. & Errors & Linguistic \\
        \midrule
        DPO & \textbf{0.7905} & \underline{0.6377} & \underline{0.0612} & \textbf{0.5405} \\
        Zero-Shot & 0.6143 & 0.6087 & 0.0408 & 0.3155 \\
        Oracle & \underline{0.6476} & \textbf{0.7101} & \textbf{0.2449} & \underline{0.4071} \\
        \bottomrule
    \end{tabular}
    \caption{Human evaluation results, where trends roughly match those on automated metrics.}
    \label{tab:turn-results-human-eval}
\end{table}

\begin{table}[h]
    \centering
    \small
    \begin{tabular}{lcccc}
        \toprule
        Pairing & Acts & Corr. & Errors & Linguistic \\
        \midrule
        Hum.-Metric & 0.7337 & 0.6891 & 0.6127 & 0.7397 \\
        Hum.-Anno. & 0.7993 & 0.7219 & -- & -- \\
        \midrule
        Hum.-Hum. & 0.4978 & 0.7187 & 0.6154 & 0.6910 \\
        \bottomrule
    \end{tabular}
    \caption{Agreement between human evaluators and automated metrics, human evaluators and LLM-assigned annotations, and inter-rater agreement.}
    \label{tab:turn-agreement}
\end{table}

\begin{table*}[t]
    \centering
    \small
    \begin{tabular}{lccccccc}
        \toprule
        Reward & Acts & Corr. & Errors & Knowledge & Cos. Sim. & ROUGE-L & Tutor Resp. \\
        \midrule
        SFT & 0.6795 & 0.5546 & 0.0506 & 0.8723 & 0.7417 & 0.3155 & 0.2102 \\
        Average & \textbf{0.6962} & \underline{0.5699} & 0.0562 & 0.8691 & 0.7433 & \textbf{0.3181} & 0.2104 \\
        \midrule
        Acts & \underline{0.6949} & 0.5677 & 0.0506 & 0.8695 & 0.7389 & 0.3111 & 0.2101 \\
        Corr. & 0.6692 & \textbf{0.5852} & 0.0506 & 0.8652 & 0.7275 & 0.3129 & 0.2081 \\
        Errors & 0.6795 & 0.5524 & 0.0562 & 0.8708 & 0.7406 & 0.3122 & 0.2105 \\
        Knowledge & 0.6846 & 0.5437 & \textbf{0.0730} & \underline{0.8756} & \textbf{0.7486} & \underline{0.3147} & \textbf{0.2124} \\
        Cos. Sim. & 0.6731 & 0.5371 & \textbf{0.0730} & 0.8713 & \underline{0.7453} & 0.2957 & \underline{0.2116} \\
        Tutor Resp. & 0.6897 & 0.5349 & \underline{0.0618} & \textbf{0.8763} & 0.7447 & 0.3137 & 0.2113 \\
        \bottomrule
    \end{tabular}
    \caption{Results of reward function ablation study. Most reward functions result in high performance on the corresponding metric. Best method is \textbf{bolded} and second best is \underline{underlined}.}
    \label{tab:turn-results-reward-ablation}
\end{table*}

\subsection{Qualitative Analysis}

We perform a qualitative analysis of model outputs to understand the strengths and weaknesses of each simulated student method.
We show example simulated utterances from each method in Table~\ref{tab:qualitative-example}, along with scores on each metric.
We show the distributions of dialogue acts and correctness for each method in Figures~\ref{fig:acts-distribution} and~\ref{fig:correctness-distribution}, respectively. We also break down the performance of each method over turns as the dialogue progresses in Figure~\ref{fig:results-turn-level}.

Generally, there are clear differences between fine-tuning and prompting methods. 
First, we find that fine-tuned models match the linguistic style of real students much more closely. They produce much shorter outputs than prompting methods (2.28 words on average for DPO 8B vs.\ 10.89 for ICL and 4.11 for real students), echoing the findings in \citet{perczel2025teachlm}. Prompting methods also consistently use typical LLM-like language features such as formal grammar and punctuation; for example, when a tutor asks ``I'll leave you to enter your answer now if you're happy with this?'', the Reasoning method generates “Yes, thanks — I'll enter D.”, while SFT generates ``yes thank you'' and the real student simply writes ``yes''.
Beyond these surface-level features, we find that fine-tuned responses follow act and correctness distributions that are much more similar to real students; prompting methods overestimate the rate of Seek Information and correct responses, and underestimate the rate of conversational acts, like Acknowledge and Off-Topic. While all methods perform poorly early in dialogues due to limited context, prompting methods suffer much more, while fine-tuning methods perform better by leveraging what they learned from real dialogues through training.

However, there are several limitations to the fine-tuning methods as well. First, they tend to write very short responses, and rarely add in typos or excessive punctuation that are found in real student responses. Moreover, all methods struggle to adapt to individual student behavior. For example, some students give very brief responses, some are very verbose, and some convey a consistent sense of helplessness; however, model response patterns tend to be uniform across students, conveying a lack of diversity. Future work can seek to address these challenges by encouraging diverse outputs in training, or by conditioning on prior dialogues or student personas if they are available, to better capture the nuance in student utterances. See Appendix~\ref{sec:additional-results} for additional discussions.

\subsection{Ablation Study}

In order to understand how each metric is related to overall student simulation quality, we run an ablation where we train DPO with rewards from only one metric at a time. We use \texttt{Llama-3.2-3B-Instruct} for this experiment and evaluate on a random 20\% split of the test data to reduce costs. We show results in Table~\ref{tab:turn-results-reward-ablation}. As expected, we see that training on a metric results in high performance on it. The exception is Errors, since the reward signal is likely too sparse to enable learning. Interestingly, training on some metrics induces very high or low performance on completely different metrics. This result may be attributed to the inherent \textit{correlation} between some metrics. For example, training on Knowledge results in the highest performance on Errors, Cos.\ Sim., and Tutor Resp., yet leads to low performance on Correctness. On the contrary, training on Correctness results in very high performance on Correctness but the lowest performance for most other metrics. 
While most relationships between aspects of student behavior are as expected, there are several surprising ones, such as between knowledge acquisition and correctness. This result reflects recent findings that training on certain human behavioral data can have unintended effects on seemingly unrelated traits~\cite{chen2025personavectorsmonitoringcontrolling}. Finally, we observe that training on the Average reward generally leads to high performance across metrics, although suffers on Knowledge, Errors, and Tutor Resp. This result suggests that while averaging is a reasonable approach, more advanced techniques for combining metrics could lead to higher overall performance.

\section{Conclusions and Future Work}

In this paper, we introduce the task of simulating student utterances in tutoring dialogues. We propose a suite of metrics that measure realism of a simulated student turn with respect to a reference student turn. We benchmark a wide range of fine-tuned and prompting-based methods, finding that while fine-tuning and RL outperform prompting, there is a long way to go before LLMs can fully resemble real student behavior in dialogues. 

There are many avenues for future work. First, since we use simple RL methods, it is possible that more advanced techniques leveraging reasoning or data augmentation would result in significantly better performance. Second, due to data constraints, we evaluate each dialogue in isolation. However, with available data, future work could include prior student information, such as prior dialogues or knowledge states, for more context in such simulations. Third, since we only explore simulation realism at the turn-level, future work should further extend our metrics to measure realism at the dialogue-level. Fourth, while we have shown that our metrics are reliable \textit{reference-based} measures of student behavior, future work should investigate \textit{reference-free} metrics for settings where ground-truth data is not available. Finally, since we only explore dialogues based on math problems, future work should extend to dialogues in other domains, such as computer science and language learning.

\section*{Acknowledgments} This work is partially supported by Renaissance Philanthropy under the learning engineering virtual institute (LEVI) initiative and the NSF under grants 2153481 and 2237676.

\section*{Limitations}

There are several practical limitations to our work. First, we only perform our experiments on a single math dialogue dataset, and it remains to be seen if our results generalize to other datasets or domains. Second, our metrics are exclusively reference-based, limiting their use to evaluation settings that are based on existing tutor-student dialogues. Third, we do not evaluate two of our metrics in the human evaluation, Knowledge Acquisition and Inducing Tutor Responses, since we believe they are highly subjective from a human perspective and therefore difficult to reliably measure in a human experiment. However, as a result, we do not have agreement between these metrics and expert ratings. Fourth, we do not measure affect or emotional state in dialogues. While student turns in our dataset give minimal signals on affect, they may be important in other domains, and should be studied in future work. Fifth, we acknowledge that our annotations and our Correctness and Errors metrics rely on proprietary LLMs. We experimented with smaller open-source LLMs in preliminary experiments and found them to be significantly less reliable. Finally, we acknowledge that RL performs only slightly better than SFT, which is perhaps surprising. While we postulate that this result is due to the inherent difficulty of the task, future work will need to develop more advanced RL methods to verify this claim.

\section*{Ethical Considerations}

There are several potential societal benefits associated with our work. Simulated students have a high potential to improve education broadly; human tutors can use them for low-stakes practice, they can be used in A/B tests to avoid risks to real students and reduce costs, and AI tutors can be trained with simulated students in the loop. Furthermore, reliable \textit{evaluations} will accelerate research on simulated students, allowing researchers to rapidly validate the strengths and weaknesses of their methods and avoid potentially deploying unreliable models in real educational settings. There are also several potential societal risks associated with our work. Simulations may be biased towards certain demographic groups due to bias in training data or inherent bias in language models. As a result, the benefits of simulated students may be diminished for students in underrepresented groups, or worse, cause unintended harm to these groups. We recommend that any simulated student methods used in real educational settings should first be evaluated across demographic groups and thoroughly evaluated for bias. Furthermore, we caution that over-reliance on simulated students could lead to a reduction in the quality of educational tools, for example, if AI tutors are trained using simulated students that do not represent real student distributions. We recommend that any educational AI tools are first thoroughly A/B tested with real students before widespread deployment to avoid harming learning outcomes for students.

\bibliography{custom}

\appendix

\section{Related Work}
\label{sec:rw}

\subsection{Student Simulation in Dialogues}

Many recent works have explored LLM-based student simulation in educational dialogues, mainly using prompting and conditioning on student personas. MathDial introduces a dialogue dataset with LLM-simulated students exhibiting specific misconceptions \cite{mathdial}, while other work uses zero-shot prompting to generate student responses for the purpose of downstream tutor training \cite{eth-tutor-rl}. Several works simulate students by conditioning LLMs on cognitive states, misconceptions, and prior knowledge, enabling diverse simulated interactions with tutoring agents \cite{liu2024socraticlm, wang-etal-2025-training, wang-etal-2024-book2dial}. Other works use LLMs to simulate student learning curves and realistic misconceptions \cite{jin2024teach, schmucker2024ruffle}, behavior that corresponds to different demographic, behavioral, or personality traits \cite{markel2023gpteach, liu-etal-2024-personality, litype}, and goals and learning trajectories \cite{2024.EDM-doctoral-consortium.122}. One recent work leverages a multi-agent LLM pipeline to refine cognitive and linguistic properties of student responses, and performs DPO on resulting data \cite{cao2026developing}. Another work leverages machine unlearning to remove knowledge of specific concepts from LLMs in order to study how they relearn these concepts through dialogue tutoring \cite{song2026simulating}. Finally, recent works have also used SFT to create simulated students, finding that this approach significantly improves authenticity \cite{perczel2025teachlm, cao2026developing}. In general, dialogue-based simulated students are typically developed for the purpose of generating data for training and evaluating LLM tutoring agents \cite{eth-tutor-rl, scarlatos2025training, wang-etal-2024-book2dial}, or creating training environments for human tutors \cite{pan2025tutorup, markel2023gpteach}.

However, these works tend to focus on simulating entire dialogues that are not grounded in real student data. In contrast, our work focuses on predicting student utterances at the turn-level, which we validate by comparing against real student data. While some prior works explicitly predict turn-level knowledge and correctness of student turns \cite{dialogue-kt,ikram-etal-2025-exploring}, they do not predict the full text of student utterances, therefore missing many other aspects of student behavior.

\subsection{Evaluating Simulated Students}

Despite growing interest in student simulation, hardly any existing work thoroughly evaluate the quality of simulated student methods and how closely do they resemble real human students. A recent survey identifies a lack of validation in many works, and proposes a ``Turing-like test'' to capture realism \cite{kaser2024simulated}. A recent work uses an LLM-conducted Turing test to measure simulated student realism, although it remains to be seen if LLMs can identify simulated students as well as humans, or if the results of the test are correlated with more reliable measures of simulation realism \cite{dou-etal-2025-simulatorarena}. A concurrent work identifies the ``competence paradox'', where  LLMs are tasked with simulating learners that are intrinsically less knowledgeable, and argues that simulated students should be constrained by definitions of student behavior and knowledge \cite{yuan2026towards}. Another recent work conducts a comprehensive analysis of simulated student realism, where expert teachers interact with simulated students, and identifies a wide range of behavioral dimensions that indicate realism and reveal weaknesses in current approaches \cite{eth-sim-student}. The identified realism indicators include error patterns, consistency in knowledge acquisition, question-asking behavior, and linguistic patterns. However, all of these indicators are evaluated using human experts, leaving the question of how to \textit{automatically} evaluate simulated students unanswered.

A common automated evaluation approach is measuring the consistency between simulated outputs and the personas they were generated on, typically via LLM-as-a-judge, used both in education \cite{liu-etal-2024-personality, wu-etal-2025-embracing, litype} and other domains \cite{abdulhai2025consistently, kyung2025patientsim, kim2025propersimdevelopingproactivepersonalized}. However, these evaluations assume access to a ground-truth persona and are often conducted in fully synthetic settings. Alternative approaches use human-evaluated realism scales \cite{mathdial, eth-sim-student, cao2026developing}, alignment between simulated performance and conditioned cognitive states \cite{smart, wang-etal-2025-training, benedetto-etal-2024-using}, accurately predicting correctness \cite{dialogue-kt}, and textual similarity between real and simulated student solutions \cite{ross2025modeling, duan2025automatedknowledgecomponentgeneration}. Finally, several works measure population-level statistics of simulated student properties, such as temporal error rates \cite{tutorgym} or length and frequency of student turns \cite{perczel2025teachlm, zhang2024simulating}, and compare against statistics measured on real student interactions.

Our evaluation metrics build on many of the ideas used in these prior works. By comparing simulated turns to ground-truth ones, we effectively measure the consistency of the simulated student. We explicitly model many of the aspects identified as important in prior works, including correctness, errors, knowledge, behavior (through dialogue acts), and similarity to real student utterances (through cosine similarity and ROUGE-L). Additionally, our work is the first to establish a framework for reference-based evaluation of simulated students in dialogues, complementing prior works that focus on fully synthetic dialogue generation and others that focus on reference-based evaluations in non-dialogue settings.

\subsection{Student Simulation in Other Settings}

Beyond dialogues, student simulation has been studied in problem-solving and assessment settings. A large body of work has studied student ``misconceptions'', particularly in math, typically fine-tuning LLMs to reliably predict errors that students will make when solving problems \cite{ross2025learningmakemistakesmodeling, fernandez-etal-2024-divert, sonkar2024llm, ross-andreas-2024-toward, daheim-etal-2024-stepwise}. Other works have fine-tuned or prompted LLMs to simulate student responses to exam questions, often for the purpose of evaluating question properties such as difficulty \cite{benedetto-etal-2024-using, smart, nguyen-etal-2025-qg, 10.1007/978-3-031-99264-3_18, he2024psychometric, 2024.EDM-long-papers.6, lu2024generative}. Another recent area of interest is studying student behavior in programming settings, typically by fine-tuning LLMs on real or simulated student traces \cite{ross2025modeling} and conditioning on student knowledge states \cite{duan2025automatedknowledgecomponentgeneration}.
Earlier non-LLM approaches modeled student problem-solving behavior and adaptation to tutor feedback in math learning platforms \cite{matsuda2015teaching}, with a focus on modeling the cognitive states of students \cite{anderson1995cognitive}.

\section{Implementation Details}
\label{sec:implementation-details}

\subsection{Models and Hyperparameters}
\label{sec:model_details}

We use \texttt{meta-llama/Llama-3.1-8B-Instruct} as the base model for all fine-tuned models, except in the experiments that use a smaller student model, where we use \texttt{meta-llama/Llama-3.2-3B-Instruct}. We load all models in floating point 16 precision. For LLM prompting, we use the \texttt{gpt-4.1-2025-04-14} version of GPT-4.1 and the \texttt{gpt-5-mini-2025-08-07} version of GPT-5 mini.

We conduct preliminary hyperparameter exploration using the validation set, optimizing for our automated metrics with the student models. We use the same hyperparameters, except for epochs (shown in Table~\ref{tab:training-stats}), across all fine-tuned models. We train using the AdamW optimizer, with learning rate = $5\cdot10^{-5}$ with a linear warmup for $10\%$ of training steps, effective batch size = $64$ via gradient accumulation, weight decay = $1\cdot10^{-2}$, and gradient norm clipping = $1.0$. We set LoRA's rank $r$ = $32$, $\alpha$ = 64, and dropout = $0.05$, with rank stabilization \cite{kalajdzievski2023rank} and adapters on all internal weight matrices. For DPO, we set the learning rate to $5\cdot10^{-6}$, $\beta=0.1$, the number of candidate student utterances per turn to $n=4$, and the reward threshold to $\epsilon=0.1$.
For DPO, we only train on a random 20\% of the dialogues in the train set for one epoch to reduce costs, and find performance to be similar to when training on the full set. In total, we form 4,998/4,703 pairs for 8B/3B DPO training, respectively. For all methods, we exclude all prompts longer than 6,000 characters at train-time to avoid memory issues.

At test-time, for all non-reasoning models, we use greedy decoding and set the maximum number of generated tokens to 400; for reasoning models, we use a temperature of $1.0$ and 15,000 maximum generated tokens. For data annotation, we use 4,000 maximum generated tokens.

In Table~\ref{tab:training-stats}, we report the validation loss, task-specific validation performance, and training/testing runtimes for each fine-tuned model used in our work. We do not include model loading in testing time, since it is a constant cost, and do not include training data evaluation in DPO training time since it is relative to the testing time of other models. We run all experiments on NVIDIA L40S and A40 GPUs, with each experiment running on a single GPU.

\begin{table*}[h]
    \centering
    \small
    \begin{tabular}{lccccc}
        \toprule
        Model & Epochs & Val. Loss & Val. Perf. & Train Time & Test Time \\
        \midrule
        \rowcolor{gray!21} \multicolumn{6}{c}{Student Models}\\
        \midrule
        Student SFT (3B) & 3 & 1.9765 & -- & 12 & 1 \\
        Student DPO (3B) & 1 & 0.6458 & -- & 56 & 1 \\
        Student SFT (8B) & 3 & 1.8761 & -- & 19 & 1 \\
        Student DPO (8B) & 1 & 0.6426 & -- & 81 & 1 \\
        \midrule
        \rowcolor{gray!21} \multicolumn{6}{c}{Metric Support Models}\\
        \midrule
        Acts (8B) & 2 & 0.0703 & 0.9219 & 150 & 1 \\
        Correctness (8B) & 1 & 0.1165 & 0.9164 & 76 & 1 \\
        Knowledge (8B) & 3 & 0.6338 & 0.6557 & 128 & 5 \\
        Tutor (8B) & 3 & 1.4802 & -- & 20 & 2 \\
        \bottomrule
    \end{tabular}
    \caption{Training statistics for all fine-tuned models. Train and test time are reported in minutes. The validation performance (Val. Perf.) metric is accuracy for Acts and Correctness and AUC for Knowledge.}
    \label{tab:training-stats}
\end{table*}

\subsection{Additional Metric Details}
\label{sec:metric-details}

\begin{table*}[htbp]
\centering
\small
\begin{tabular}{p{2.5cm}p{12cm}}
\toprule
Dialogue Act  & Description \\
\midrule
Math Answer  & When the tutor asks a math content-related question, the student attempts to answer that question. \\
Not Understanding  & The student simply indicates that they do not know the answer to a question or do not understand a concept.  \\
Seek Information & The student seeks more information regarding the math problem or topic, for example, by asking a clarifying or conceptual question. \\
Off-Topic & The student utterance is unrelated to the problem or math topic, including greetings, goodbyes, and other casual conversation. \\
Acknowledge & The student simply acknowledges what the tutor said in the previous turn. \\
\bottomrule
\end{tabular}
\caption{Dialogue acts that students can make at any turn in a dialogue.}
\label{tab:acts}
\end{table*}

We show our dialogue act definitions in Table~\ref{tab:acts}. To derive our set of acts, we started with definitions from prior work \cite{dialogue-acts, 10.1007/978-3-031-98417-4_9, studychat} and then refined them to fit our setting of one-on-one math tutoring and to retain a level of granularity that balanced low ambiguity and high behavioral descriptiveness. We then internally labeled acts and compared with LLM labels. Afterward, we further refined our definitions to address areas of disagreement and confusion. For example, we originally included an ``Other'' act, but found that we could achieve better labeling accuracy by explicitly labeling each utterance’s act. We also originally included distinct acts for correct and incorrect math answers, but collapsed those to avoid overlapping with the correctness metric.

For the acts and correctness models, we train using a simple SFT objective where the output is the text of the corresponding label, and use greedy decoding for inference. We clarify that the correctness model is only used for the analysis in Figure~\ref{fig:correctness-distribution}, and not the correctness metric which uses LLM prompting. We implement our KT model using the LLMKT implementation from~\cite{dialogue-kt}. Additionally, we condition the KT model on estimated OCEAN personas, as detailed in Section~\ref{sec:benchmarks}, which increases AUC significantly from 0.5940 to 0.6557. We note that this modification does not leak information to the student model at test-time since the KT model is only used for evaluation. We also acknowledge that the performance of the KT model may appear low; however, it is a difficult task since correctness is being predicted without observing the question the tutor asks. Furthermore, the goal of the KT model is to learn a correlation between utterances and future correctness, forming a complement to the correctness metric that compares correctness for the current turn.

\subsection{Software}

We use the Azure API for prompting GPT-4.1 for data annotation and the OpenAI API for prompting GPT-4.1 and GPT-5-mini for all other purposes\footnote{\url{https://platform.openai.com/docs/libraries}}. We use the sentence transformers library~\cite{reimers-2019-sentence-bert} for computing sentence embeddings. We load Llama models and perform SFT using the Huggingface Transformers library~\cite{wolf-etal-2020-transformers}, perform DPO using the trl library~\cite{vonwerra2022trl}, and perform LoRA using the peft library~\cite{peft}. We perform local inference using vLLM~\cite{kwon2023efficient}. We perform all other standard machine learning operations using PyTorch~\cite{paszke2019pytorch} and numpy~\cite{harris2020array}, perform data loading and transformation using pandas~\cite{mckinney-proc-scipy-2010}, compute statistics using SciPy~\cite{2020SciPy-NMeth} and scikit-learn~\cite{scikit-learn}, and compute ROUGE-L using rouge-score\footnote{\url{https://github.com/google-research/google-research/tree/master/rouge}}. Our work complies with the terms of use for all software we use.

\section{Additional Dataset Details}
\label{sec:dataset-details}

We now provide additional details on the Question-Anchored Tutoring Dialogues 2k dataset. The dataset is completely anonymized \cite{eedi-dialogues}, and as a result there is no demographic information available on individual students or tutors. All students are from the United Kingdom. The dataset is licensed using cc-by-nc-sa 4.0, and we release our annotations under the same license.

Dialogues are 23.42 turns long on average across student and tutor turns, with a minimum length of 10 and a maximum of 109. Dialogues can be initiated by tutors or students, with 82.63\% of dialogues initiated by tutors. Student/tutor turns are 4.11/14.84 words long on average, respectively, with both containing a mixture of English, numbers, mathematical symbols, and emojis. We show the distribution of act labels in Figure~\ref{fig:acts-distribution} and the distribution of correctness labels in Figure~\ref{fig:correctness-distribution}.

Since solutions are not provided in the dataset, we use GPT-4.1 to annotate each question with the correct answer, a textual solution, and an explanation for each multiple-choice option. We also ask the model to identify if the question is solvable, since several questions are associated with images that were not properly translated to text or contain other data processing issues. We exclude such unsolvable questions from the dataset, resulting in 60 dialogues (3\% of the total) being removed.

Each dialogue is associated with a set of ``subjects'', which we use to form the set of knowledge components (KCs) $C$ for our knowledge acquisition metric. We only use the most granular level of subject definitions (level 3 in the dataset).  We also include an additional ``Default'' KC for each dialogue that the LLM can assign to turns that may not fall into the given subjects. Not including Default, there are 157 unique subjects in the dataset, with each dialogue being associated with an average of 4.57 subjects.

In rare cases, GPT-4.1 fails to annotate dialogues with acts or correctness. Act annotation failures occur 2/1/1 times in the train/validation/test splits, respectively, and correctness annotation failures occur 1/2/2 times in the train/validation/test splits, respectively. In these cases, we simply do not include turns in failed dialogues when computing metrics that rely on the ground-truth labels, at both train and test time.

\section{Additional Results}
\label{sec:additional-results}

\subsection{Additional Qualitative Analysis}

We continue our discussion of qualitative findings to identify areas for improvement in fine-tuned methods. In addition to generating overly short responses, fine-tuned methods also significantly underestimate the Seek Information act, and are less likely than real students to ask questions, identified as common behavioral inconsistencies in LLMs~\cite{eth-sim-student}. We hypothesize that these differences may be due to using greedy decoding for generation; since a majority of real student responses are three or less words long (62\%), the \textit{most likely} response at any point in time will likely be very short. The under-representation of Seek Information further indicates that the models have not learned the nuances of when to use rare linguistic features or generate longer responses that are associated with the act. While random sampling can produce longer and more diverse responses, we found that it leads to significantly reduced performance in preliminary experiments. These findings indicate that more advanced training or decoding techniques, such as overgenerate-and-rank or MCTS, may be required to obtain longer and more complex responses while maintaining accuracy on realism metrics.

\subsection{Label Distributions}

\begin{figure*}[htbp]
    \centering
    \includegraphics[width=1\linewidth]{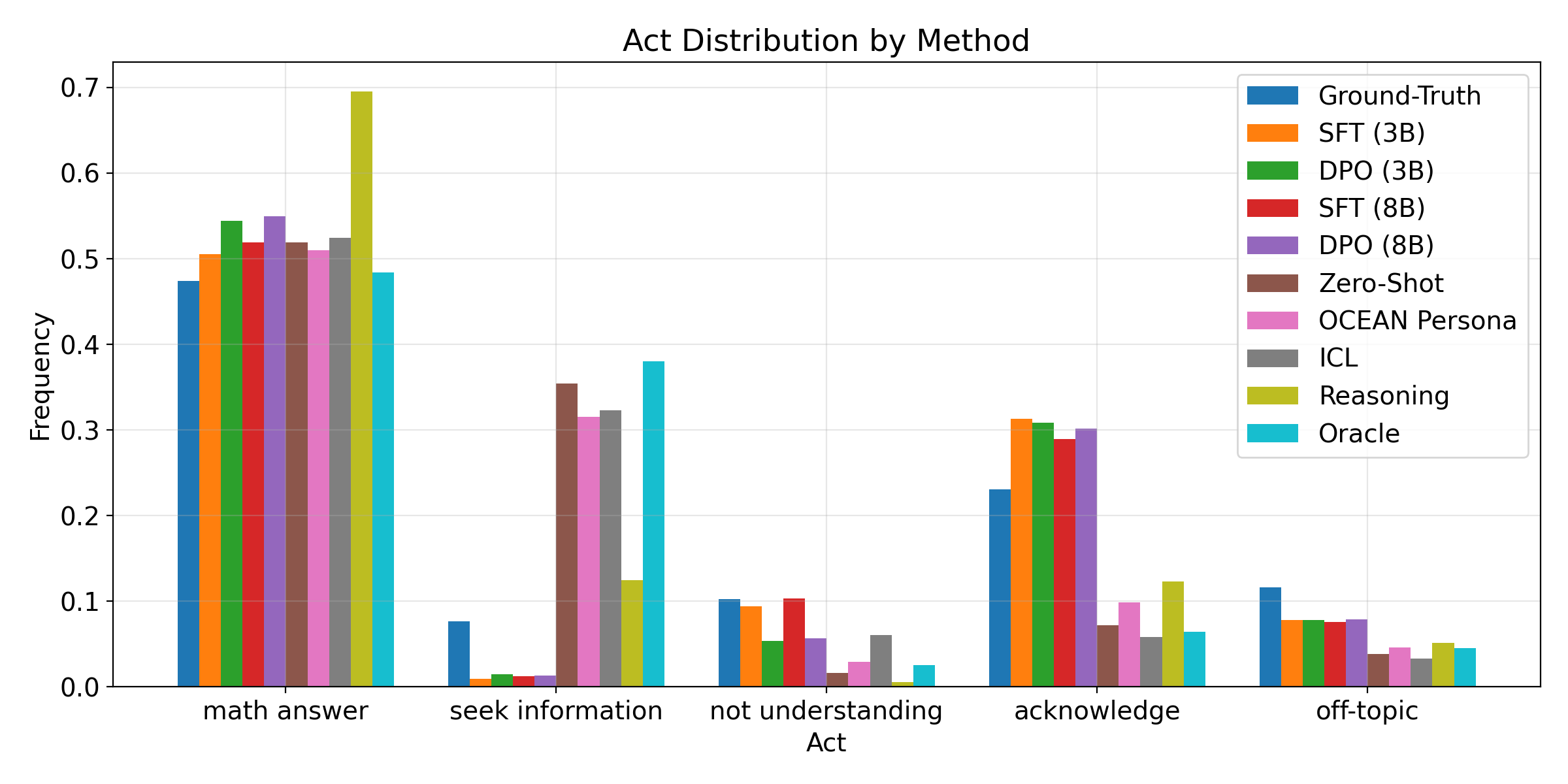}
    \caption{Distribution of act labels for real students and simulated student methods.}
    \label{fig:acts-distribution}
\end{figure*}

\begin{figure*}[htbp]
    \centering
    \includegraphics[width=1\linewidth]{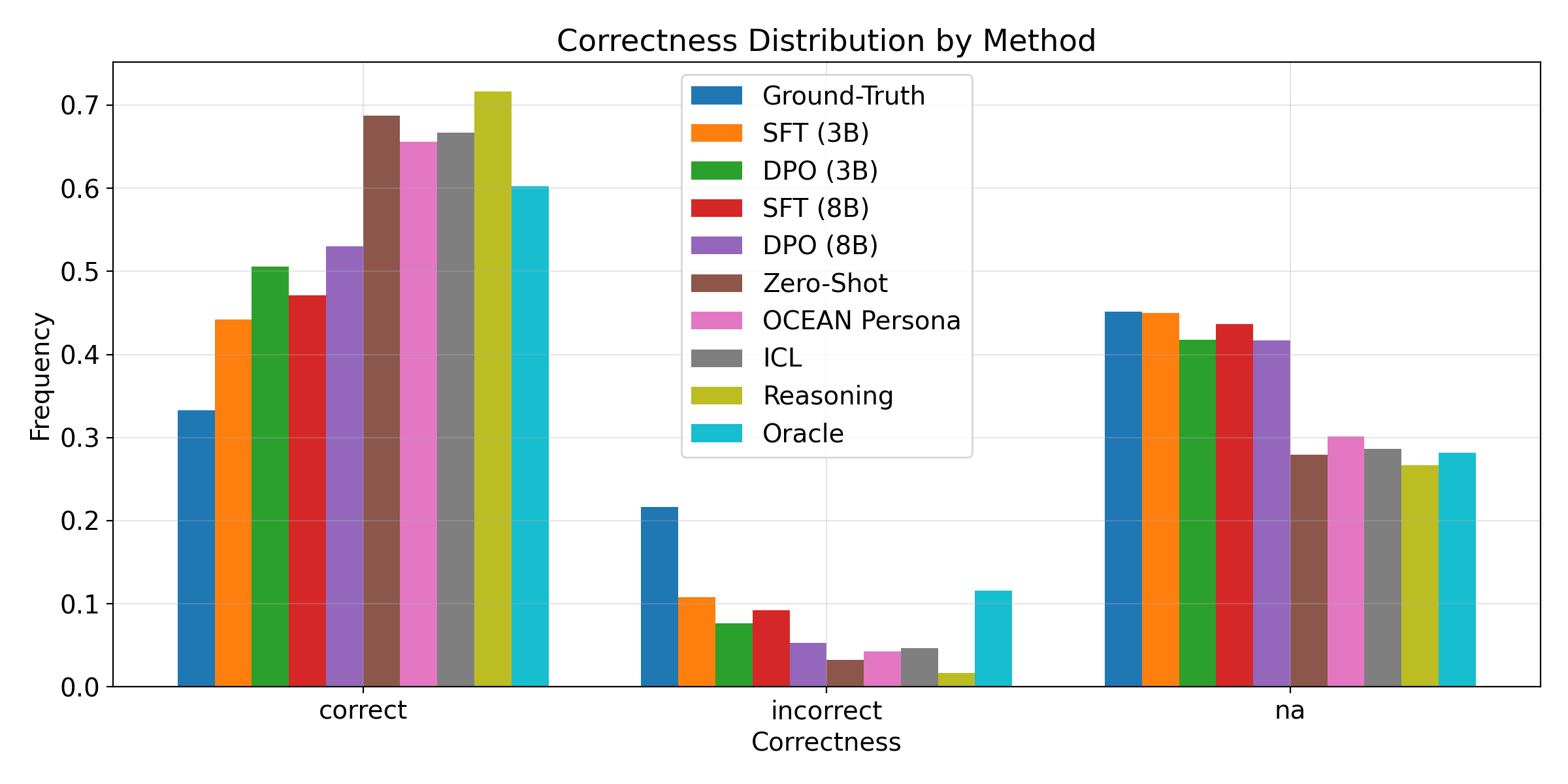}
    \caption{Distribution of correctness labels for real students and simulated student methods.}
    \label{fig:correctness-distribution}
\end{figure*}

We examine the distribution of both acts and correctness to better understand the behavior of different student models. Using the full test set, for each student model, we compute the portion of turns that are classified for each possible act and correctness label, using fine-tuned act and correctness classifiers, respectively. We compare these to the distribution of acts and correctness for the ground-truth, real student turns on the test set.

We show the distribution of acts across methods in Figure~\ref{fig:acts-distribution}. We observe that while most methods have similar frequencies of Math Answer, the distributions for other acts vary greatly. Compared to prompting-based methods, the fine-tuned methods have rates that are much more similar to the ground-truth for Not Understanding, Acknowledge, and Off-Topic. However, fine-tuned methods rarely use the Seek Information act. A possible reason for this is that the fine-tuned responses tend to be very short, following common linguistic patterns in the data, while Seek Information turns tend to be longer since they involve asking specific questions. On the other hand, prompting-based methods overly use Seek Information, with lower rates for Not Understanding, Acknowledge, and Off-topic. This shows that prompting-based methods are reluctant to generate more conversational turns that are common in student dialogues, while overestimating how often students ask questions. The Reasoning model stands out because it explicitly reasons about what acts to take, but still is less accurate than the fine-tuned models.

We show the distribution of correctness across methods in Figure~\ref{fig:correctness-distribution}. We observe that fine-tuned methods are much closer to the ground-truth distribution than prompting methods, with prompting methods more frequently giving correct answers and less frequently giving answers that do not have a particular correctness ($na$). This result explains how prompting-based methods perform better on the correctness metric, because they are more likely to predict the majority class. However, both fine-tuned and prompting-based methods overestimate how often students are correct, demonstrating a need for methods to better anticipate when students will give incorrect answers.

\subsection{Results by Turn}

\begin{figure*}[htbp]
    \centering
    \includegraphics[width=1\linewidth]{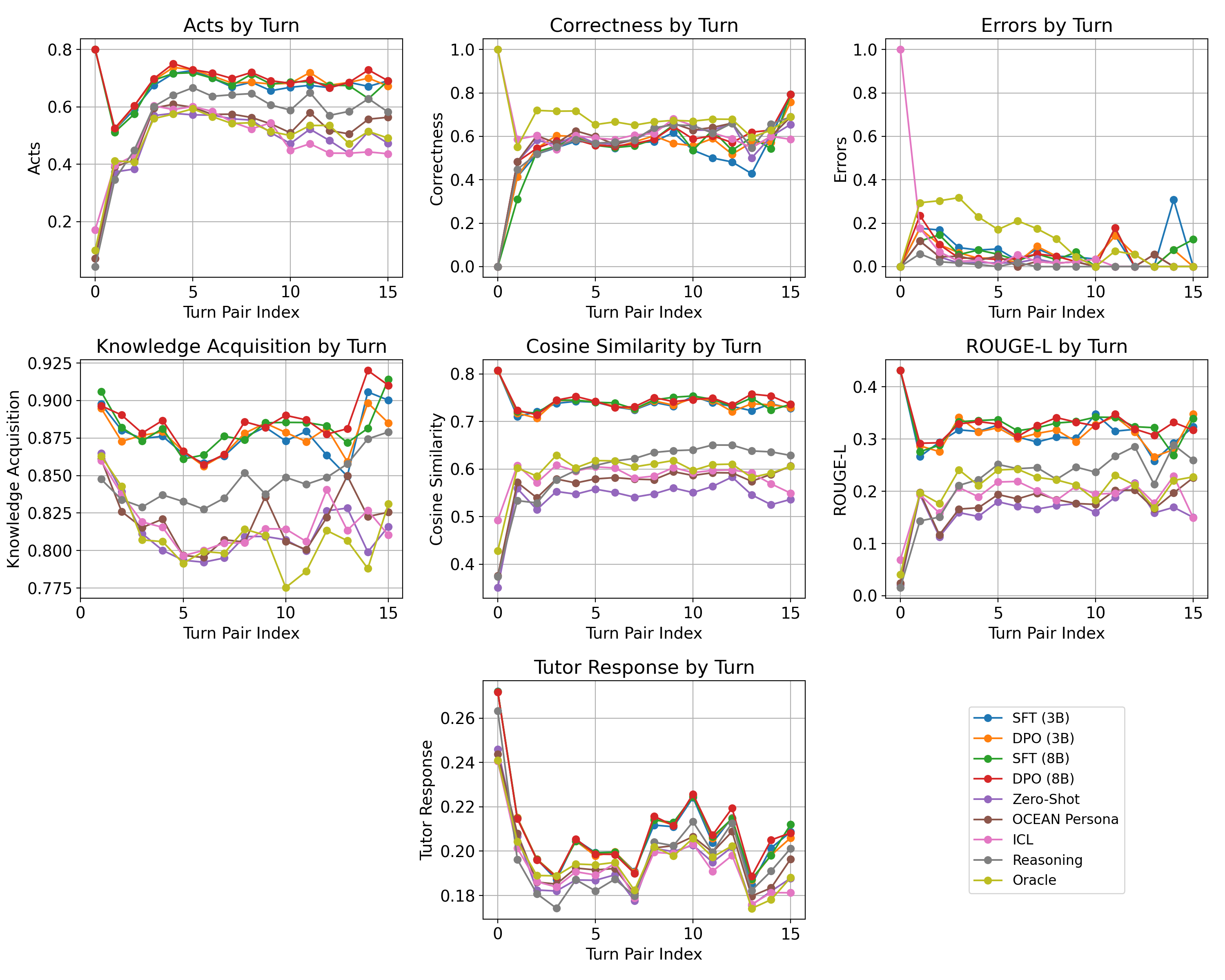}
    \caption{Results across metrics and methods broken down for each turn pair index.}
    \label{fig:results-turn-level}
\end{figure*}

To investigate how challenging different aspects of student simulation are at different points in dialogues, we examine how well methods perform on each of our metrics at the turn level. We plot the average at each turn pair index across the test set. To exclude high variance results, we truncate all dialogues after 15 turn pairs; 83\% of dialogues are within this length.

We show the results of this investigation in Figure~\ref{fig:results-turn-level}. We observe that all metrics show variability across the turns of a dialogue. Many metrics, such as Acts, Cosine Similarity, ROUGE-L, and Tutor Response, are very easy for fine-tuned methods on the first turn due to most dialogues starting with a simple greeting. Prompting methods, on the other hand, perform poorly on the first turn since they have no prior turns in context. Correctness and Errors vary greatly on the first turn since most first turns have $na$ as their correctness; in the rare cases where the first turn is correct or incorrect, some methods like Oracle correctly predict this, while others are not able to without context. After the first turn, the metrics are mostly stable, with a few exceptions: Acts are more challenging early on, since it is hard to predict the student's behavior with limited context. On the contrary, Knowledge Acquisition, becomes more difficult between turns 0 and 6, since the problem solving process begins in this range and it is difficult to estimate student knowledge with limited context; afterward, as context increases, so does Knowledge Acquisition performance. Fine-tuned methods tend to dominate prompting methods across turns, although the difference is usually greater early in the dialogues, since the prompting methods perform much better when there is more context. We also observe that the Reasoning method dominates other prompting methods later in the dialogues on Acts, Knowledge Acquisition, Cosine Similarity, and ROUGE-L, showing that it benefits from the added context more than other methods.

\section{Human Evaluation Details}
\label{sec:human_eval_details}

In this section, we detail the instructions, recruitment details, and the annotation interface for our human evaluation. The study was approved by the Institutional Review Board (IRB).

\begin{figure*}[htbp]
    \centering
    \begin{minipage}{\textwidth}
        \centering
        \includegraphics[width=1\linewidth]{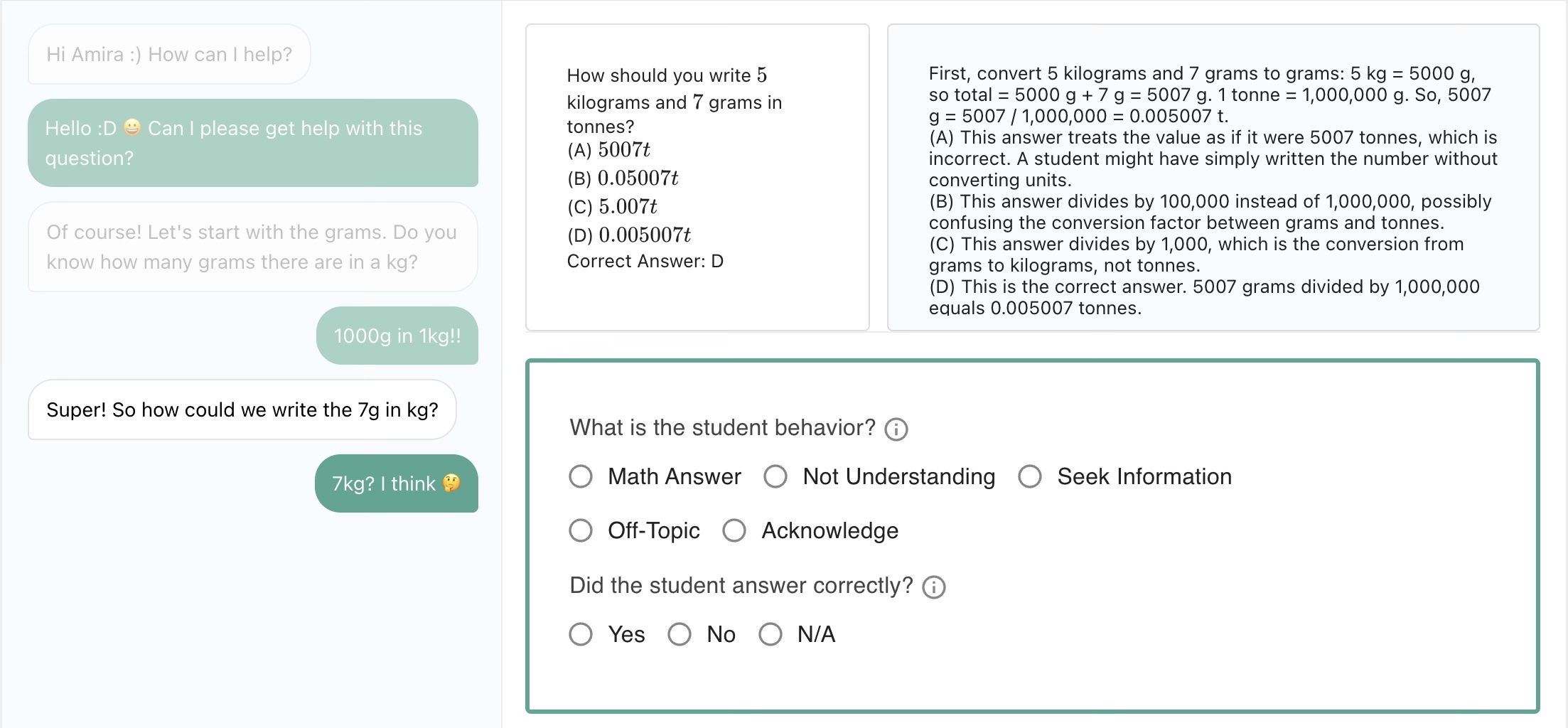}
        \caption{Human evaluation interface for evaluating ground-truth turns.}
        \label{fig:task-a-gt}
    \end{minipage}

    \vspace{0.8em}

    \begin{minipage}{\textwidth}
        \centering
        \includegraphics[width=1\linewidth]{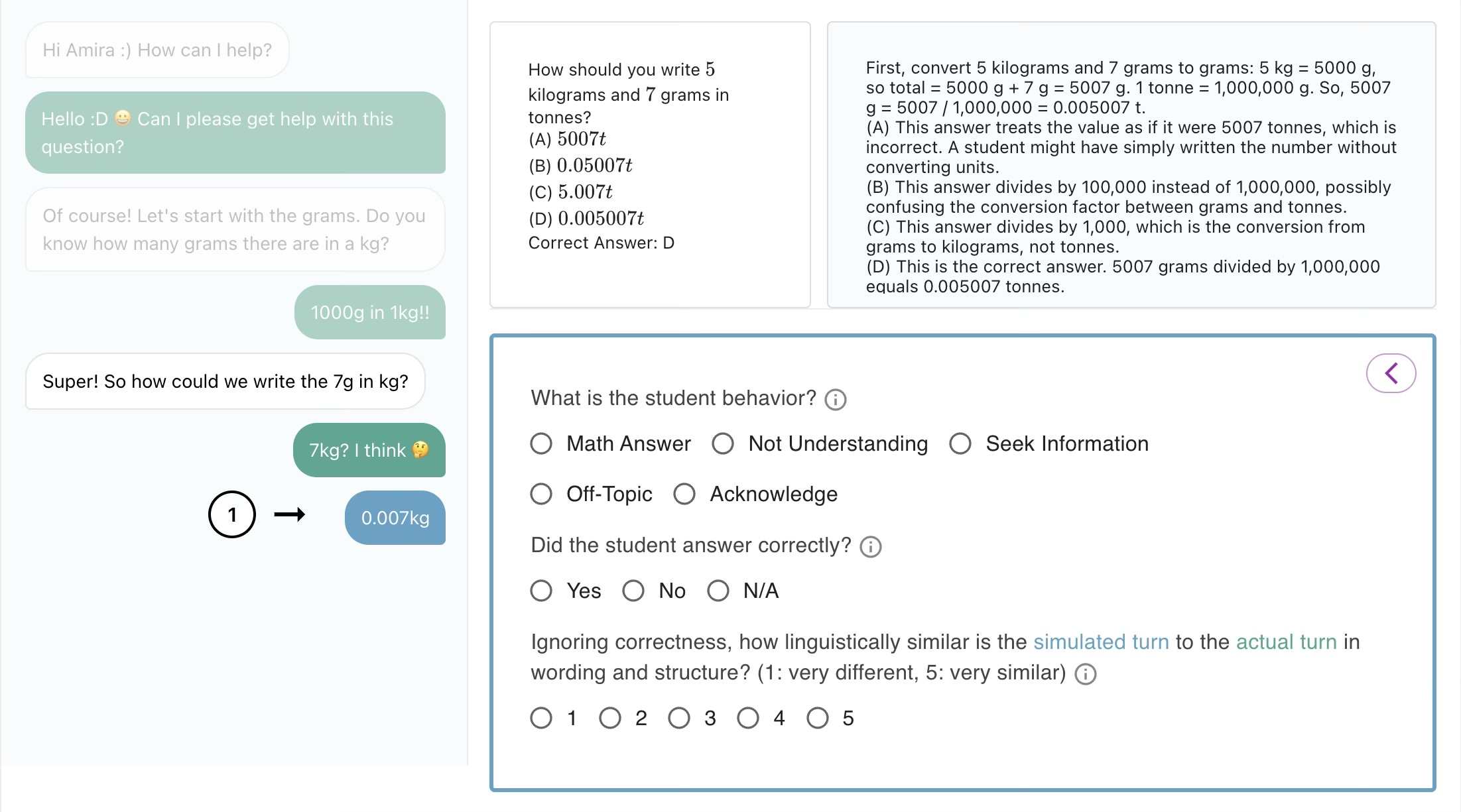}
        \caption{Human evaluation interface for evaluating simulated turns.}
        \label{fig:task-a-pred}
    \end{minipage}
\end{figure*}

Table~\ref{tab:job_description} presents the job posting used to recruit participants on Upwork. We hired four annotators and excluded the data from one due to consistently low-quality submissions. All annotators were paid \$60 USD for 2 hours of work, which is significantly higher than the minimum wage in the United States. Instructions were provided in the form of slides, with separate slides corresponding to each annotation table, including Table~\ref{tab:acts}, Table~\ref{tab:correctness}, and Table~\ref{tab:linguistic_similarity}, as well as screenshots of the annotation environment shown in Figure~\ref{fig:task-a-gt} and Figure~\ref{fig:task-a-pred}. Table~\ref{tab:consent_form} contains the consent form used in the study.

\begin{table*}[t]
\centering
\small
\begin{tabular}{|p{0.95\textwidth}|}
\hline
We are seeking a qualified math teacher or tutor to evaluate AI-generated dialogues centered on middle school math problems. The ideal candidate will have experience in mathematics education and the ability to assess both the realism of student dialog turns. Responsibilities include reviewing dialogue content and providing ratings based on specific evaluation metrics. If you are passionate about education and interested in shaping how AI can support learning, we’d love to hear from you! For detail, please read the following post:

---

\textbf{Help Shape the Future of AI Tutoring — \$60 for \textasciitilde2 Hours (Remote, Flexible)}

Are you a math teacher or tutor in the US or UK? Your expertise can make AI tutoring actually useful for real learners. Lend your judgment, get paid, and help set the standard for quality in educational AI.\\
\\
\textbf{What you’ll do}

* Review short or full student–tutor dialogues tied to middle-school math problems\\
* Rate how realistic and helpful the AI’s responses are\\
* No prep, no grading, no lesson plans—just your professional judgment\\
\\
\textbf{Why join}

* Real impact: Your ratings directly influence how AI supports students\\
* Simple + flexible: Complete online on your schedule in one sitting\\
* Fair pay: \$60 for a single \textasciitilde2-hour session (paid upon full completion)\\
\\
\textbf{Who we’re looking for}

* Adults (18+) who read and understand English\\
* Prior math teaching or tutoring experience in the United States or United Kingdom (required)\\
\\
\textbf{How it works}

* You’ll be doing a turn-level evaluation\\
* You’ll complete multiple problems within that single task\\
* Clear instructions provided; straightforward rating interface\\
\\
\textbf{Confidentiality}

* Contact/consent info (name, email) collected only for study admin and deleted after analysis\\
* Data are anonymized and securely stored; any quotes used will not include identifying details\\
\\
\textbf{Interested? Apply here on Upwork}
\\
\hline
\end{tabular}
\caption{Recruitment posting for math teachers and tutors for human evaluation.}
\label{tab:job_description}
\end{table*}

\newpage
\onecolumn

\begin{small}
\begin{longtable}{|p{0.95\textwidth}|}
\hline
\textbf{Study Invitation}

You are invited to participate in a research study evaluating an AI-generated student dialogue for math problems. This study is conducted by a research group from \textbf{[University Name]}. \\
\\
\textbf{Why are we doing this research study?}

The purpose of this research study is to better understand how people evaluate AI-generated tutoring dialogues. We aim to identify which types of AI responses seem most similar to what real students would say. By studying how participants judge individual dialogue turns, we can develop clearer criteria for evaluating educational AI systems and improve their reliability and usefulness for learners. \\
\\
\textbf{Who can participate in this research study?}

Adults aged 18 years or older who can read and understand English and who have prior mathematics teaching or tutoring experience in the United States or the United Kingdom are eligible to participate in this study. \\
\\
\textbf{What will I be asked to do and how much time will it take?}

Participants will read several middle-school--level math problems and the accompanying dialogue materials. Participants will evaluate individual AI-generated responses within a dialogue produced by different AI systems. For each item, participants will provide ratings describing how appropriate or realistic the AI responses are. All math problems are at the middle-school level, such as ratio-chaining items (for example, ``a:b = 3:2 and b:c = 3:4; what is a:c?''). 
\\
\begin{itemize}[nosep]
    \item Participants will complete multiple problems within their assigned task. The total time required for the study is approximately two hours. Participants may work at their own pace but must complete their assigned task in full to receive compensation. 
    \item Participants will first read a description of the study and then provide informed consent before beginning any study activities. 
    \item Participants will complete several math problems during the study. For each problem, they will begin by reading and understanding the problem before proceeding to the corresponding dialogue materials. 
    \item Participants will read a short dialogue associated with each math problem. After reading the context, they will review several AI-generated candidate responses representing possible student turns at specific moments in the dialogue. For each candidate, participants will rate how similar the response is to what a student would plausibly say at that point. This sequence will repeat for each assigned problem. 
\end{itemize}

\\
\textbf{Will being in this research study help me in any way?}

Being in this research study is not expected to provide any direct personal benefit to you. However, your participation may help researchers better understand how people evaluate AI-generated tutoring dialogues and may contribute to the improvement of future educational technologies. \\
\\
\textbf{What are my risks of being in this research study?}

There is always a risk of breach of confidentiality, but this has been mitigated as described below. The risks involved in participating in this study are minimal, similar to what you might experience during a typical teaching activity. You are free to take breaks or stop at any time if you wish. \\
\\
\textbf{How will my personal information be protected?}

Your participation will be kept confidential. We will collect your name, phone number, and email for contact and consent purposes, but this information will be deleted after the experiment analysis is complete. All personal information will be anonymized, and any quotes presented in a paper will not include identifying details. The analyzed data will be securely stored in anonymized form, with access restricted to the research team. \\
\\
\textbf{Will I be given any money or other compensation for being in this research study?}

Yes. Participants will receive monetary compensation for completing the study. Compensation is provided at a rate of \$30 per hour, for a total of \$60 for completing the two-hour session. Participants must complete their assigned task in full to receive compensation. \\
\\
\textbf{What happens if I say yes, but I change my mind later?}

Participation in this study is completely voluntary. You may withdraw at any time without any penalty or loss of benefits. If you decide to withdraw, any data collected up to that point will be used anonymously unless you request otherwise. \\
\\
\textbf{Who can I talk to if I have questions?}

If you have questions about this project or decide to drop out, you may contact the research team at \textbf{[Researcher Email]}. If you have questions concerning your rights as a research subject, you may contact the \textbf{[University Human Research Protection Office]} at \textbf{[HRPO Phone Number]} or \textbf{[HRPO Email]}. \\
\\
\textbf{Consent}

Please sign (or e-sign) below to indicate your consent, then scan or photograph this page and submit it as instructed by the study team. \\
\\
\textbf{Printed Name:} \rule{0.6\textwidth}{0.4pt} \\
\\
\textbf{Date:} \rule{0.4\textwidth}{0.4pt} \\
\\
Please print or download a copy of this page for your records.\\
\hline
\caption{Informed consent form for our human evaluation.}
\label{tab:consent_form}
\end{longtable}
\end{small}
\clearpage
\twocolumn

\begin{table*}
\centering
\small
\begin{tabular}{lp{10cm}}
\toprule
Correctness  & Description \\
\midrule
Yes  & Student correctly responds to the previous tutor turn. \\
No  & Student incorrectly responds to the previous tutor turn, or indicates they do not know the answer.  \\
N/A & All other cases, such as when the tutor does not ask a question or only asks a conversational question, or if the student response is purely conversational. A turn is conversational when it does not address a mathematical task posed by the tutor. \\
\bottomrule
\end{tabular}
\captionof{table}{Correctness, \textit{Did the student answer correctly?}}
\label{tab:correctness}
\end{table*}

\begin{table*}
\centering
\small
\begin{tabular}{p{1cm}p{5cm}p{8.5cm}}
\toprule
Rating  & Similarity & Description \\
\midrule
1 & Not linguistically similar & Completely different wording, structure, or response type. \\ 
2 & Slightly linguistically similar & Very limited overlap (e.g., both contain a number or keyword), but overall expression is different. \\
3 & Moderately linguistically similar & Same general type of response, but noticeable differences in wording, structure, or detail. \\
4 & Highly linguistically similar & Very similar wording and structure, with minor differences. \\
5 & Nearly identical linguistically & Almost the same wording, structure, and level of detail. \\
\bottomrule
\end{tabular}
\captionof{table}{Linguistic Similarity, \textit{Ignoring correctness, how linguistically similar is the simulated turn to the actual turn in wording and structure?}}
\label{tab:linguistic_similarity}
\end{table*}

\onecolumn

\section{Examples}

\begin{table*}[h]
    \centering
    \small
    \begin{tabular}{lccccccc}
        \toprule
        \multicolumn{8}{c}{Question}\\
        \midrule
        \multicolumn{8}{p{.8\textwidth}}{Alex has two fraction cards. \( \frac{5}{8} \quad \square\) The product of the cards is \( \frac{1}{6} \) What is the other card?\newline
        A: \( \frac{19}{24} \)\newline
        B: \( \frac{5}{48} \)\newline
        C: \( \frac{15}{4} \)\newline
        D: \( \frac{4}{15} \)\newline
        Correct Answer: D}\\
        \midrule
        \multicolumn{8}{c}{Dialogue Context}\\
        \midrule
        \multicolumn{8}{p{.8\textwidth}}{
        Tutor: Hello again Leo! How can I help?\newline
        Student: I'm don't get the fact about missing number questions.\newline
        Tutor: Well, we know that 5/8 x ? = 1/6. So we need to rearrange that to get ? =\newline
        Student: Would we do the inverse.\newline
        Tutor: What would that be?\newline
        \fcolorbox{green}{yellow!0}{Student: 5/8 divided by 1/6?}\newline
        \fcolorbox{yellow}{yellow!0}{SFT 3B: 8/5}\newline
        \fcolorbox{yellow}{yellow!0}{DPO 3B: 8/5}\newline
        \fcolorbox{yellow}{yellow!0}{SFT 8B: 5/8}\newline
        \fcolorbox{yellow}{yellow!0}{DPO 8B: 5/8 divided by 1/6}\newline
        \fcolorbox{yellow}{yellow!0}{Zero-Shot: So, would I divide 1/6 by 5/8?}\newline
        \fcolorbox{yellow}{yellow!0}{OCEAN: Um, so would I divide 1/6 by 5/8? I think that's the inverse, right?}\newline
        \fcolorbox{yellow}{yellow!0}{ICL: So like, 1/6 divided by 5/8?}\newline
        \fcolorbox{yellow}{yellow!0}{Reasoning: Yes — multiply by the reciprocal. x = (1/6) ÷ (5/8) = (1/6)*(8/5) = 8/30 = 4/15}\newline
        \fcolorbox{yellow}{yellow!0}{Oracle: So, like, the opposite of multiplying? So we would divide 1/6 by 5/8? Is that right?}\newline
        Tutor: Nearly! We need to divide the left by 5/8 to get ? on its own, so we must do the same to the right hand side}\\
        \midrule
        \multicolumn{8}{c}{Next Student Turn Scores}\\
        Source & Acts (Label) & Corr. (Label) & Errors & Knowledge (Quant.) & Cos. Sim. & ROUGE-L & Tutor Resp. \\
        \midrule
        Ground-Truth & -- (MA) & -- (incorrect) & -- & -- (4.00) & -- & -- & -- \\
        SFT 3B & 1 (MA) & 0 (correct) & 0 & 0.9167 (3.67) & 0.5388 & 0.2500 & 0.1893 \\
        DPO 3B & 1 (MA) & 0 (correct) & 0 & 0.9167 (3.67) & 0.5388 & 0.2500 & 0.1893 \\
        SFT 8B & 1 (MA) & 1 (incorrect) & 0 & 0.9167 (3.67) & 0.6600 & 0.5000 & 0.1818 \\
        DPO 8B & 1 (MA) & 1 (incorrect) & 1 & 0.9167 (3.67) & 0.9567 & 1.0000 & 0.1818 \\
        Zero-Shot & 0 (SI) & 0 (correct) & 0 & 0.7500 (3.00) & 0.8714 & 0.2667 &  0.1818\\
        OCEAN & 0 (SI) & 0 (correct) & 0 & 0.9167 (3.67) & 0.7874 & 0.1739 & 0.1798 \\
        ICL & 1 (MA) & 0 (correct) & 0 & 0.8333 (3.33) & 0.9039 & 0.2857 & 0.1720 \\
        Reasoning & 1 (MA) & 0 (correct) & 0 & 1.0000 (4.00) & 0.7960 & 0.3333 & 0.1531 \\
        Oracle & 0 (SI) & 0 (correct) & 0 & 0.7500 (3.00) & 0.7498 & 0.1667 & 0.1778 \\
        \bottomrule
    \end{tabular}
    \caption{Qualitative example for a single turn, showing the ground-truth student response (bordered in green), the simulated responses for that turn from all methods (bordered in yellow), and corresponding scores across all metrics. The ground-truth student response is an incorrect math answer (MA) to the previous tutor turn, where the correct answer would divide in the other direction. Most simulated responses are classified as MA, while some are classified as Seek Information (SI) due to having a more questioning tone. All simulated responses except for SFT 8B and DPO 8B are correct, with SFT 3B and DPO 3B inverting 5/8, most prompting methods inverting the full equation, and Reasoning giving a complete solution to the problem. While SFT 8B and DPO 8B are both incorrect, only DPO 8B has the same error as the ground-truth student response. For knowledge, we show the score and the average delta quantile across KCs. In this example, knowledge is highly correlated with task difficulty, reflecting findings in~\cite{dialogue-kt}. Specifically, incorrect or partially correct answers are likely to lead to easier followup tutor questions due to scaffolding, and are therefore more likely to be answered correctly by the student. On the other hand, correct answers are likely to be followed by the tutor posing a new task, which is likely to be harder than a scaffolded question, leading to a lower likelihood of the student answering correctly. The knowledge increases for Reasoning because the question is solved, so no new tasks will be posed. Cos. Sim. and ROUGE-L are highly correlated with textual overlap. Finally, Tutor Resp. is lower for prompting methods since they give full solutions, which would be unlikely to lead to the ground-truth tutor response.}
    \label{tab:qualitative-example}
\end{table*}

\begin{table*}[h]
    \centering
    \small
    \begin{tabular}{llcc}
        \toprule
        \multicolumn{4}{c}{Question}\\
        \midrule
        \multicolumn{4}{p{.8\textwidth}}{$ \frac{A}{10}=\frac{9}{15} $ What is the value of $ A $ ?\newline
        A: $3$\newline
        B: $4$\newline
        C: $6$\newline
        D: $9$\newline
        Correct Answer: C}\\
        \midrule
        \multicolumn{4}{c}{Knowledge Components (KCs)}\\
        \midrule
        \multicolumn{4}{p{.8\textwidth}}{Fractions of an Amount\newline
        Mental Addition and Subtraction\newline
        Equivalent Fractions\newline
        Dividing Fractions\newline
        Naming Co-ordinates in 2D\newline
        Default}\\
        \midrule
        \multicolumn{4}{c}{Dialogue Context}\\
        \midrule
        \multicolumn{4}{p{.8\textwidth}}{
        Tutor: Have you tried simplifying 9/15?\newline
        Student: yes. 3/5\newline
        Tutor: Excellent! Good work!!!! Now how would you get that to a denominator of 10?\newline
        Student: what do you mean. 3/5\newline
        Tutor: so in 3/5 you have 5 as the denominator\newline
        Student: oh\newline
        Tutor: this question needs 10\newline
        Student: oh you times the numerator and denominator by 2\newline
        Tutor: Exactly!!!!! So what do you think it will become?}\\
        \midrule
        \multicolumn{4}{c}{Next Student Turn}\\
        Source & Utterance & $\nabla Z$ & $\operatorname{quant}(\nabla Z)$\\
        \midrule
        Ground-Truth & c & [0.0117, 0.0117, -0.0156, 0.0039, 0.0078, -0.0078] & [2, 2, 1, 2, 2, 1]\\
        Simulated & 6/10 & [0.0391, 0.0352, 0.0391, 0.0352, 0.0156, 0.0352] & [3, 3, 3, 3, 3, 3]\\
        \midrule
        \multicolumn{4}{c}{Delta Quantile Upper Bounds: [-0.0234, 0.0000, 0.0156, 0.0430, $\infty$]}\\
        \midrule
        \multicolumn{4}{c}{Knowledge Acquisition Similarity}\\
        \midrule
        \multicolumn{4}{c}{$1 - \frac{|2-3| + |2-3| + |1-3| + |2-3| + |2-3| + |1-3|}{4 \times 6}=0.6666$}\\
        \bottomrule
    \end{tabular}
    \caption{Example decomposition of Knowledge Acquisition similarity for a single turn. The simulated turn is from DPO 8B. Early turns in the dialogue with greetings are excluded for brevity. While both answers are correct, the ground-truth simply gives the final answer option, while the simulated turn gives the full fraction associated with the correct answer. The knowledge tracing model estimates that the student's knowledge delta for the ground-truth turn is roughly average for all KCs (2 or 1 between 0 and 4), while the knowledge delta for the simulated turn is above average for all KCs (all 3). This difference may be due to the possibility that answering ``c'' is simply a lucky guess, whereas answering ``6/10'' conveys an understanding of the solution, leading to a higher knowledge estimate. This example demonstrates the need for a knowledge-based metric, since two turns can have the same correctness but give different indications of student mastery on concepts.}
    \label{tab:knowledge-breakdown}
\end{table*}

\begin{small}
\begin{longtable}{p{\textwidth}}
        \toprule
        \multicolumn{1}{c}{Question}\\
        \midrule
        Arrange the numbers in descending order. \( -5,-8,7,-6 \)\newline
        A: \( -8,7,-6,-5 \)\newline
        B: \( 7,-8,-6,-5 \)\newline
        C: \( -8,-6,-5,7 \)\newline
        D: \( 7,-5,-6,-8 \)\newline
        Correct Answer: D\\
        \midrule
        \multicolumn{1}{c}{Question Solution (Generated by GPT-4.1)}\\
        \midrule
        Solution: To arrange the numbers -5, -8, 7, -6 in descending order, we start with the largest and go to the smallest. 7 is the largest, then -5, then -6, then -8. So the correct order is 7, -5, -6, -8.\newline
        A Explanation: This option lists -8 first, which is the smallest number, not the largest. This is the reverse of the correct order (ascending order for the negatives, but with 7 misplaced). A student might mistakenly think that more negative numbers are larger.\newline
        B Explanation: This option starts with 7, which is correct, but then puts -8 before -6 and -5. This is incorrect because -8 is the smallest (most negative) number. A student might have placed 7 first but then listed the negatives in the wrong order.\newline
        C Explanation: This option lists all the negative numbers first in increasing order, then 7 at the end. This is ascending order, not descending. A student might have confused ascending and descending order.\newline
        D Explanation: This is the correct answer: 7, -5, -6, -8. 7 is the largest, followed by -5, then -6, then -8.\\
        \midrule
        \multicolumn{1}{c}{Dialogue}\\
        \midrule
        Tutor: Hi again! Firstly, do you know what descending order is?\newline
        Student: YEARH!\newline
        Tutor: Go on, what does that mean?\newline
        Student: Biggest to smallest\newline
        Tutor: Spot on!! From those number which number would be the biggest?\newline
        Student: \color{green}\ding{51}\color{black}. Wich numbers\newline
        Tutor: We have -5 -8 7 -6\newline
        Student: Ohh -5\newline
        Tutor: 7 must be the biggest number as it is the only positive number. Try and imagine or even draw out a number line, 7 is the only number greater than 0. So from our answer options we can narrow it down do B or D as they both start with 7. Does that make sense so far?\newline
        Student: \color{green}\ding{51}\color{black}\newline
        Tutor: Excellent\newline
        Student: \color{green}\ding{51}\ding{51}\ding{51}\ding{51}\ding{51}\ding{51}\ding{51}\ding{51}\ding{51}\ding{51}\ding{51}\ding{51}\ding{51}\ding{51}\ding{51}\ding{51}\ding{51}\ding{51}\ding{51}\ding{51}\ding{51}\ding{51}\ding{51}\ding{51}\ding{51}\ding{51}\color{black}\newline
        Tutor: Now we have these number left -5 -8 -6\newline
        Student: \color{green}\ding{51}\color{black}\newline
        Tutor: From these number which one is the biggest? So try and think of a number line, which number is closest to 0?\newline
        Student: -5\newline
        Tutor: Well done Mariana, that points us to answer D! 7, -5, -6, -8\newline
        Student: D\color{green}\ding{51}\color{black}\newline
        Tutor: Bye for now!\newline
        Student: Bye thanks for your help!\\
        \midrule
        \multicolumn{1}{c}{OCEAN Persona (Generated by GPT-4.1)}\\
        \midrule
        Reasoning: The student demonstrates understanding of mathematical concepts when prompted and is able to follow the tutor's guidance. Their responses are brief and sometimes enthusiastic (e.g., 'YEARH!', multiple checkmarks), but do not show much elaboration or curiosity beyond the immediate task, suggesting neutral openness. The student is responsive and follows instructions, indicating a reasonable level of conscientiousness, though there is no evidence of independent organization or planning, so this is also neutral. The student is not very talkative and gives short answers, which suggests low extraversion. The student is polite, cooperative, and expresses gratitude at the end, indicating high agreeableness. There is no evidence of anxiety, frustration, or negative emotions, so neuroticism appears low.\newline
        Openness: neutral\newline
        Conscientiousness: neutral\newline
        Extraversion: low\newline
        Agreeableness: high\newline
        Neuroticism: low\\
        \midrule
        \multicolumn{1}{c}{Oracle Summary (Generated by GPT-4.1)}\\
        \midrule
        The student demonstrates a basic understanding of mathematical concepts when prompted, correctly identifying "descending order" as "biggest to smallest" and, with guidance, selecting the correct answer. Initially, the student makes a conceptual error by choosing -5 as the largest number, indicating some confusion about the relative size of negative and positive numbers. However, the student is receptive to feedback, quickly adjusts their thinking when the tutor explains, and successfully applies the number line strategy when prompted. Behaviorally, the student is highly responsive, frequently uses affirmative emojis (\color{green}\ding{51}\color{black}) and enthusiastic language, and often provides brief, sometimes minimal, answers, suggesting a preference for quick confirmation over detailed explanation. The student is agreeable and polite, expressing gratitude at the end, and shows openness to learning by engaging with the tutor's questions. Linguistically, the student uses informal, enthusiastic responses, sometimes with spelling errors ("YEARH!," "Wich numbers," "Ohh -5"), and tends to favor emoji-heavy, concise replies over elaboration. Overall, the student is eager, agreeable, and responsive, with a tendency toward impulsive answers but a willingness to learn and correct mistakes when guided.\\
        \midrule
        \multicolumn{1}{c}{Retrieved ICL Example}\\
        \midrule
        Student: Hello!\newline
        Tutor: Hiya! How can I help?\newline
        Student: May you help me please? Am very stuck!\newline
        Tutor: of course!\newline
        Student: Thank you\newline
        Tutor: are you stuck on writing numbers in desending order?\newline
        Student: yes\newline
        Tutor: do you know what descending means?\newline
        Student: Is it largest to smallest? I don’t really know?\newline
        Tutor: that's correct!\newline
        Student: Oh…\newline
        Tutor: so which one is the largest?\newline
        Student: 7\newline
        Tutor: good, now which is the next biggest?\newline
        Student: -5\newline
        Tutor: good! now what?\newline
        Student: -6\newline
        Tutor: good! and finally?\newline
        Student: -8\newline
        Tutor: good!\newline
        Student: oh that was easy! hehe\newline
        Tutor: would you like any more help?\newline
        Student: i want to say a big thank you for helping me!\newline
        Tutor: hahah no problem! You did very well! Have a nice evening!\\
        \bottomrule
    \caption{Example dialogue, question, solution, and associated context provided to OCEAN, Oracle, and ICL prompting methods. Emojis except checkmark are removed.}
    \label{tab:persona-examples}
\end{longtable}
\end{small}

\clearpage

\section{Prompts}
\label{sec:prompts}

\subsection{Annotations}

\begin{table*}[h]
\centering
\small
\begin{promptbox}
\begin{tabular}{p{\linewidth}}
You are a math education expert. Your job is to label the **dialogue acts** for student turns in a given dialogue. \\[0.5em]
These are the available dialogue act labels: \\
- Math Answer: When the tutor asks a math content-related question, the student attempts to answer to that question \\
- Seek Information: The student seeks more information regarding the math problem or topic, for example, by asking a clarifying or conceptual question \\
- Not Understanding: The student simply indicates that they do not know the answer to a question or do not understand a concept \\
- Acknowledge: The student simply acknowledges what the tutor said in the previous turn \\
- Off-Topic: The student utterance is unrelated to the problem or math topic, including greetings, goodbyes, and other casual conversation \\[0.5em]
For each **student turn** in the dialogue, choose the dialogue act that best describes the turn. Pick exactly one act for each turn from the list above, and write the dialogue act name exactly as it appears. Before writing the acts for a turn, provide reasoning about what the best act should be. \\[0.5em]
Please provide your answer as a JSON object with the following format: \\
\{ \\
\quad "turn n": \{ \\
\quad\quad "reasoning": "...", \\
\quad\quad "act": "..." \\
\quad \}, \\
\quad "turn n+2": \{ \\
\quad\quad "reasoning": "...", \\
\quad\quad "act": "..." \\
\quad \}, \\
\quad ... \\
\} \\
\end{tabular}
\end{promptbox}
\caption{System prompt for annotating dialogue acts.}
\end{table*}

\begin{table*}[h]
\centering
\small
\begin{promptbox}
\begin{tabular}{p{\linewidth}}
You are an experienced math teacher and education expert. You are given a dialogue between a student and tutor where the student is trying to solve a math problem. Your job is to identify when the student responds correctly to the tutor. Please follow these instructions carefully when making your prediction: \\
- For each student turn, identify the correctness of the student's response to the previous tutor turn. \\
- Correctness can be true, false, or null. It is true when the student correctly responds to the previous tutor turn. It is false if the student incorrectly responds to the previous tutor turn, or indicates they do not know the answer. It is null in all other cases, such as when the tutor does not ask a question or only asks a conversational question, or if the student response is purely conversational. A turn is conversational when it does not address a mathematical task posed by the tutor. \\
- Before making each correctness prediction, write a short summary of each student turn in the dialogue. The summary should include the task previously posed by the tutor, and explain why the student's response is correct, incorrect, or conversational. \\
- Your final prediction should be a JSON object using the template: \{"turn n": \{"summary": ..., "correct": true/false/null\}, "turn n+2": ...\}. \\
- Use the turn index from the conversation history as the key in your result. There should be exactly one entry for each student turn in the dialogue. \\
\end{tabular}
\end{promptbox}
\caption{System prompt for annotating correctness.}
\end{table*}

\begin{table*}[h]
\centering
\small
\begin{promptbox}
\begin{tabular}{p{\linewidth}}
You are an experienced math teacher and education expert. You are given a dialogue between a student and tutor where the student is trying to solve a math problem. Your job is to list the knowledge components (KCs) that can be used to classify the learning objectives at each turn in this dialogue. Please follow these instructions carefully when making your prediction: \\
- Tutor turns are often phrased as questions or tasks. In these cases, choose KCs that the student will need in order to respond correctly to the tutor's question. If the tutor turn does not pose a question or task, then you do not need to assign KCs to it. \\
- You will be given a list of KCs to choose from. When choosing them, write them exactly as they appear. \\
- If the tutor posed a task but none of the given KCs apply, assign "Default". \\
- Write a short summary of each tutor turn in the dialogue, including the intended learning objectives. \\
- Along with each summary, list ALL candidate KCs that can be used to describe each tutor turn in the dialogue. \\
- Your final response should be a JSON object using the template: \{"turn n": \{"summary": "...", "kcs": ["kc 1 id", "kc 2 id", ...]\}, "turn n+2": ...\} \\
- Use the turn index from the conversation history as the key in your result. There should be exactly one entry for each tutor turn in the dialogue. \\
\end{tabular}
\end{promptbox}
\caption{System prompt for annotating knowledge components.}
\end{table*}

\begin{table*}[h]
\centering
\small
\begin{promptbox}
\begin{tabular}{p{\linewidth}}
You are a math education expert. Your task is to analyze the options of math multiple choice questions. Follow these instructions carefully: \\
- First attempt to solve the problem. If it is not possible to solve the problem because it is poorly defined, then say the problem is not solvable. \\
- Then write an explanation for each option. If the option is the correct answer, write the correct solution to reach that answer. If the option is an incorrect answer, explain the error a student might make to reach that answer. \\
- Give your final response as a JSON object with the following template: \\
\{ \\
\ \ "solution": ..., \\
\ \ "solvable": true/false, \\
\ \ "correct\_option": 1-4, \\
\ \ "option\_1\_explanation": ..., \\
\ \ "option\_2\_explanation": ..., \\
\ \ "option\_3\_explanation": ..., \\
\ \ "option\_4\_explanation": ... \\
\} \\
\end{tabular}
\end{promptbox}
\caption{System prompt for annotating question solutions.}
\end{table*}

\begin{table*}[h]
\centering
\small
\begin{promptbox}
\begin{tabular}{p{\linewidth}}
You are analyzing a dialogue between a student and a math tutor. Your task is to assess the student's personality based on the OCEAN model, also known as the Big Five Traits.  \\

**OCEAN Traits Description:** \\
- **Openness to Experience:** Reflects the student's curiosity, creativity, willingness to try new things, and openness to new ideas and experiences.  \\
- **Conscientiousness:** Indicates the student's level of organization, diligence, responsibility, and reliability in approaching tasks.  \\
- **Extraversion:** Represents how outgoing, energetic, and socially confident the student appears.  \\
- **Agreeableness:** Measures the student's friendliness, cooperativeness, compassion, and willingness to collaborate.  \\
- **Neuroticism:** Assesses the student's emotional stability, tendency to experience negative emotions such as anxiety, moodiness, or vulnerability to stress. \\

First provide reasoning about the student's behavior with respect to the OCEAN model. Then, determine if the student's expression of each trait is **high**, **neutral**, or **low**. Base your reasoning only on the dialogue provided. In your final answer, output your results as a JSON object with the following template: \\
\{ \\
\ \ "reasoning": "...", \\
\ \ "Openness": "low/neutral/high", \\
\ \ "Conscientiousness": "low/neutral/high", \\
\ \ "Extraversion": "low/neutral/high", \\
\ \ "Agreeableness": "low/neutral/high", \\
\ \ "Neuroticism": "low/neutral/high" \\
\} \\
\end{tabular}
\end{promptbox}
\caption{System prompt for annotating OCEAN personas.}
\end{table*}

\begin{table*}[h]
\centering
\small
\begin{promptbox}
\begin{tabular}{p{\linewidth}}
You are analyzing a dialogue between a student and a math tutor. Your task is to summarize the student's persona based on their interactions in the dialogue. Focus on the following aspects: \\
- How well the student acquires knowledge during the dialogue. \\
- The types of mathematical errors the student makes. \\
- Any notable behavioral patterns, such as frequent question asking, immediately jumping to the answer, distracting from the task at hand, etc. \\
- The student's personality traits, such as openness, conscientiousness, extraversion, agreeableness, and neuroticism. \\
- Notable linguistic patterns in the student's responses. \\
\\
Your response should be a single paragraph summarizing the student's persona. \\
\end{tabular}
\end{promptbox}
\caption{System prompt for creating Oracle summaries.}
\end{table*}

\clearpage
\subsection{Evaluation Models}

\begin{table*}[h]
\centering
\small
\begin{promptbox}
\begin{tabular}{p{\linewidth}}
Your task is to classify the dialogue acts for the last turn in the given dialogue.
\end{tabular}
\end{promptbox}
\caption{System prompt for fine-tuned act classifier model.}
\end{table*}

\begin{table*}[h]
\centering
\small
\begin{promptbox}
\begin{tabular}{p{\linewidth}}
Your task is to classify whether the last student turn in the given dialogue is one of: "correct", "incorrect", or "na".
\end{tabular}
\end{promptbox}
\caption{System prompt for fine-tuned correctness classifier model.}
\end{table*}

\begin{table*}[h]
\centering
\small
\begin{promptbox}
\begin{tabular}{p{\linewidth}}
You are a tutor guiding a student through a math problem.
\end{tabular}
\end{promptbox}
\caption{System prompt for fine-tuned tutor model.}
\end{table*}

\begin{table*}[h]
\centering
\small
\begin{promptbox}
\begin{tabular}{p{\linewidth}}
You are a math education expert. You will observe a tutoring dialogue where a student is attempting to solve a math problem. You will see two versions of the next student turn: a ground-truth turn and a candidate turn. Your job is to evaluate the correctness and errors of the candidate turn.\newline
- The correctness of the ground-truth turn is given. You must evaluate the correctness of the candidate turn.\newline
- Correctness can be "correct", "incorrect", or "na". It is "correct" if the student correctly responds to the previous tutor turn. It is "incorrect" if the student incorrectly responds to the previous tutor turn, or indicates they do not know the answer. It is "na" in all other cases, such as when the tutor does not ask a question or only asks a conversational question, or if the student response is purely conversational. A turn is conversational when it does not address a mathematical task posed by the tutor.\newline
- If both the ground-truth AND candidate turns are "incorrect", evaluate if they have the same error. They have the same error if the two turns are mathematically EQUIVALENT. If they are mathematically inequivalent, they do NOT have the same error.\newline
\newline
After reasoning, please return the correctness of the **candidate** turn as "correct", "incorrect", or "na". If both the ground-truth and candidate turns are incorrect, add "same error" or "different error" to your response (ex: "incorrect, same error"). Do not include any other text in your response.
\end{tabular}
\end{promptbox}
\caption{System prompt for correctness and error prediction.}
\end{table*}

\begin{table*}[h]
\centering
\small
\begin{promptbox}
\begin{tabular}{p{\linewidth}}
You are an experienced math teacher. You are given a dialogue between a student and teacher where a student is trying to solve a math problem. Your job is to predict if the student has a particular knowledge component (KC) at the current point in the dialogue. Please follow these instructions carefully when making your prediction:\newline
- The student will need to possess this KC in order to respond correctly to the teacher's most recent question.\newline
- Use previous information in the dialogue to determine if the student has this KC or not.\newline
- Only respond with a single word, "True" or "False".\newline
\end{tabular}
\end{promptbox}
\caption{System prompt for knowledge tracing model.}
\end{table*}

\clearpage
\subsection{Student Models}

\begin{table*}[h]
\centering
\small
\begin{promptbox}
\begin{tabular}{p{\linewidth}}
You are a student attempting to solve a math problem, seeking help from a tutor.
\end{tabular}
\end{promptbox}
\caption{System prompt for fine-tuned student model.}
\end{table*}

\begin{table*}[h]
\centering
\small
\begin{promptbox}
\begin{tabular}{p{\linewidth}}
You will act as a student in a conversation with a teacher in training. You will need to act as much like a student as possible. If possible do not respond with overly long messages. The conversation with the teacher will be about the following math problem. You may or may not know how to solve it already, let the teacher guide you to the correct understanding. You will be tested at the end and scored thus it is best if you collaborate with the teacher as it has more experience in math than you. If you believe you have figured out the problem and don't need any more help, put <end\_of\_dialogue> after your response.
\end{tabular}
\end{promptbox}
\caption{System prompt for Zero-Shot student model.}
\end{table*}

\begin{table*}[h]
\centering
\small
\begin{promptbox}
\begin{tabular}{p{\linewidth}}
You will act as a student in a conversation with a teacher in training. You will need to act as much like a student as possible. If possible do not respond with overly long messages. The conversation with the teacher will be about the following math problem. You may or may not know how to solve it already, let the teacher guide you to the correct understanding. You will be tested at the end and scored thus it is best if you collaborate with the teacher as it has more experience in math than you. If you believe you have figured out the problem and don't need any more help, put <end\_of\_dialogue> after your response.\newline\newline
You will be given a Big Five persona that describes how you should act in the dialogue. Follow this persona as closely as possible.
\end{tabular}
\end{promptbox}
\caption{System prompt for OCEAN student model.}
\end{table*}

\begin{table*}[h]
\centering
\small
\begin{promptbox}
\begin{tabular}{p{\linewidth}}
You will act as a student in a conversation with a teacher in training. You will need to act as much like a student as possible. If possible do not respond with overly long messages. The conversation with the teacher will be about the following math problem. You may or may not know how to solve it already, let the teacher guide you to the correct understanding. You will be tested at the end and scored thus it is best if you collaborate with the teacher as it has more experience in math than you. If you believe you have figured out the problem and don't need any more help, put <end\_of\_dialogue> after your response.\newline\newline
You will be given a persona that describes how you should act in the dialogue. Follow this persona as closely as possible.
\end{tabular}
\end{promptbox}
\caption{System prompt for Oracle student model.}
\end{table*}

\begin{table*}[h]
\centering
\small
\begin{promptbox}
\begin{tabular}{p{\linewidth}}
You will act as a student in a conversation with a teacher in training. You will need to act as much like a student as possible. If possible do not respond with overly long messages. The conversation with the teacher will be about the following math problem. You may or may not know how to solve it already, let the teacher guide you to the correct understanding. You will be tested at the end and scored thus it is best if you collaborate with the teacher as it has more experience in math than you. If you believe you have figured out the problem and don't need any more help, put <end\_of\_dialogue> after your response.\newline\newline
You will also be given an example of a previous dialogue. Your responses should be similar to the ones in this example.
\end{tabular}
\end{promptbox}
\caption{System prompt for ICL student model.}
\end{table*}

\begin{table*}[h]
\centering
\small
\begin{promptbox}
\begin{tabular}{p{\linewidth}}
You will act as a student in a conversation with a teacher in training. You will need to act as much like a student as possible. If possible do not respond with overly long messages. The conversation with the teacher will be about the following math problem. You may or may not know how to solve it already, let the teacher guide you to the correct understanding. You will be tested at the end and scored thus it is best if you collaborate with the teacher as it has more experience in math than you. If you believe you have figured out the problem and don't need any more help, put <end\_of\_dialogue> after your response.\newline\newline
Your response will be judged on how well it matches what the actual student said next in the dialogue (unseen). The following criteria will be used to evaluate your response:\newline
- Acts: Does your response make the same dialogue act as the real student response\newline
- Correctness: Does your response have the same correctness as the real student response\newline
- Errors: If your response is an incorrect math answer, does it have the same underlying error as the real student response\newline
- Knowledge: Does your response represent the same mastery of knowledge concepts as the real student response\newline
- Linguistic: Does your response have the same linguistic features as the real student response\newline
\newline
These are the available dialogue acts:\newline
- Math Answer: When the tutor asks a math content-related question, the student attempts to answer to that question\newline
- Seek Information: The student seeks more information regarding the math problem or topic, for example, by asking a clarifying or conceptual question\newline
- Not Understanding: The student simply indicates that they do not know the answer to a question or do not understand a concept\newline
- Acknowledge: The student simply acknowledges what the tutor said in the previous turn\newline
- Off-Topic: The student utterance is unrelated to the problem or math topic, including greetings, goodbyes, and other casual conversation\newline
\newline
These are the available correctness states:\newline
- Correct: The student correctly responds to the mathematical task posed in the previous tutor turn\newline
- Incorrect: The student incorrectly responds to the mathematical task posed in the previous tutor turn or indicates they don't know the answer\newline
- NA: The tutor doesn't pose a task that has a clear correct/incorrect answer OR the student doesn't indicate correctness in their response\newline
\newline
Reason about how to respond in order to maximize the evaluation criteria. Your final response should only contain the predicted student utterance.\newline
\end{tabular}
\end{promptbox}
\caption{System prompt for Reasoning student model.}
\end{table*}

\end{document}